\documentclass{article}

\usepackage{changepage}

\usepackage[utf8]{inputenc}

\usepackage{textcomp,marvosym}

\usepackage{fixltx2e}

\usepackage{amsmath,amssymb}

\usepackage{cite}

\usepackage{hyperref}

\usepackage[right]{lineno}

\usepackage{rotating}

\usepackage{color}

\usepackage{url}

\usepackage[labelfont=bf]{caption}

\makeatletter
\renewcommand{\@biblabel}[1]{\quad#1.}
\makeatother

\date{}

\usepackage{lastpage,fancyhdr,graphicx}

\graphicspath{{./Figures/}}
\newcommand{\rmd}{\mathrm{d}}

\newcommand{\mL}{\mathcal{L}}

\newcommand{\bx}{\mathbf{x}}
\newcommand{\bR}{\mathbf{R}}
\newcommand{\be}{\mathbf{e}}

\newcommand{\by}{\mathbf{y}}
\newcommand{\bn}{\mathbf{n}}

\newcommand{\quotpo}{\mathbb{R}^3 \rtimes S^2}
\newcommand{\FBC}{\text{FBC}}
\newcommand{\RFBC}{\text{RFBC}}
\newcommand{\LFBC}{\text{LFBC}}
\newcommand{\AFBC}{\text{AFBC}}
\begin{document}

\title{Improving Fiber Alignment in HARDI by Combining Contextual PDE Flow with Constrained Spherical Deconvolution}
\author{J.M. Portegies\textsuperscript{1},
R.H.J. Fick\textsuperscript{2},
G.R. Sanguinetti\textsuperscript{1},\\
S.P.L. Meesters\textsuperscript{1,3},
G. Girard\textsuperscript{2,4},
R. Duits\textsuperscript{1,5}}
\maketitle
\begin{flushleft}
\bf{1} Dept. of Math. and Computer Science, TU/e,  The Netherlands\\
\bf{2} Athena, INRIA Sophia Antipolis, France
\\
\bf{3} Academic Center for Epileptology Kempenhaeghe \& Maastricht UMC+, The Netherlands
\\
\bf{4} SCIL, Universit\'e de Sherbrooke, Canada
\\
\bf{5} Dept. of Biomedical Eng., TU/e, The Netherlands

* E-mail: j.m.portegies@tue.nl
\end{flushleft}

\section*{Abstract}
We propose two strategies to improve the quality of tractography results computed from diffusion weighted magnetic resonance imaging (DW-MRI) data. Both methods are based on the same PDE framework, defined in the coupled space of positions and orientations, associated with a stochastic process describing the enhancement of elongated structures while preserving crossing structures. In the first method we use the  enhancement PDE for contextual regularization of a fiber orientation distribution (FOD) that is obtained on individual voxels from high angular resolution diffusion imaging (HARDI) data via constrained spherical deconvolution (CSD). Thereby we improve the FOD as input for subsequent tractography. Secondly, we introduce the fiber to bundle coherence (FBC), a measure for quantification of fiber alignment. The FBC is computed from a tractography result using the same PDE framework and provides a criterion for removing the spurious fibers. We validate the proposed combination of CSD and enhancement on phantom data and on human data, acquired with different scanning protocols. On the phantom data we find that PDE enhancements improve both local metrics and global metrics of tractography results, compared to CSD without enhancements. On the human data we show that the enhancements allow for a better reconstruction of crossing fiber bundles and they reduce the variability of the tractography output with respect to the acquisition parameters. Finally, we show that both the enhancement of the FODs and the use of the FBC measure on the tractography improve the stability with respect to different stochastic realizations of probabilistic tractography. This is shown in a clinical application: the reconstruction of the optic radiation for epilepsy surgery planning.

\section{Introduction}

 Diffusion weighted magnetic resonance imaging (DW-MRI) is a non-invasive technique for the characterization of biological tissue microstructure \cite{leBihan1986}. In brain white matter, water molecules diffuse predominantly along axonal fibers. This results in an observable macroscopic orientation dependence in the DW signal, that is measured by scanning the tissue in multiple orientations and gradient strengths. To model the angular anistropy of the diffusion profile, diffusion tensor imaging (DTI) \cite{Basser1994} is widely used, but this has the limitation that only a single fiber direction can be estimated per voxel \cite{tournier2011diffusion}. It is estimated in \cite{jeurissen2013investigating} that more complex fiber configurations occur in approximately 90\% of the white matter voxels. To overcome this, high angular resolution diffusion imaging (HARDI) techniques are used, that can describe more complex (crossing) fiber configurations. An overview of HARDI techniques can be found in \cite{Descoteaux2012}. Here we use the method of constrained spherical deconvolution (CSD) \cite{Tournier2012}, that from the initial diffusion data constructs a fiber orientation distribution (FOD), which models the distribution of fibers along different directions.

\begin{figure}[t!]
  \centering
  \includegraphics[width=\textwidth]{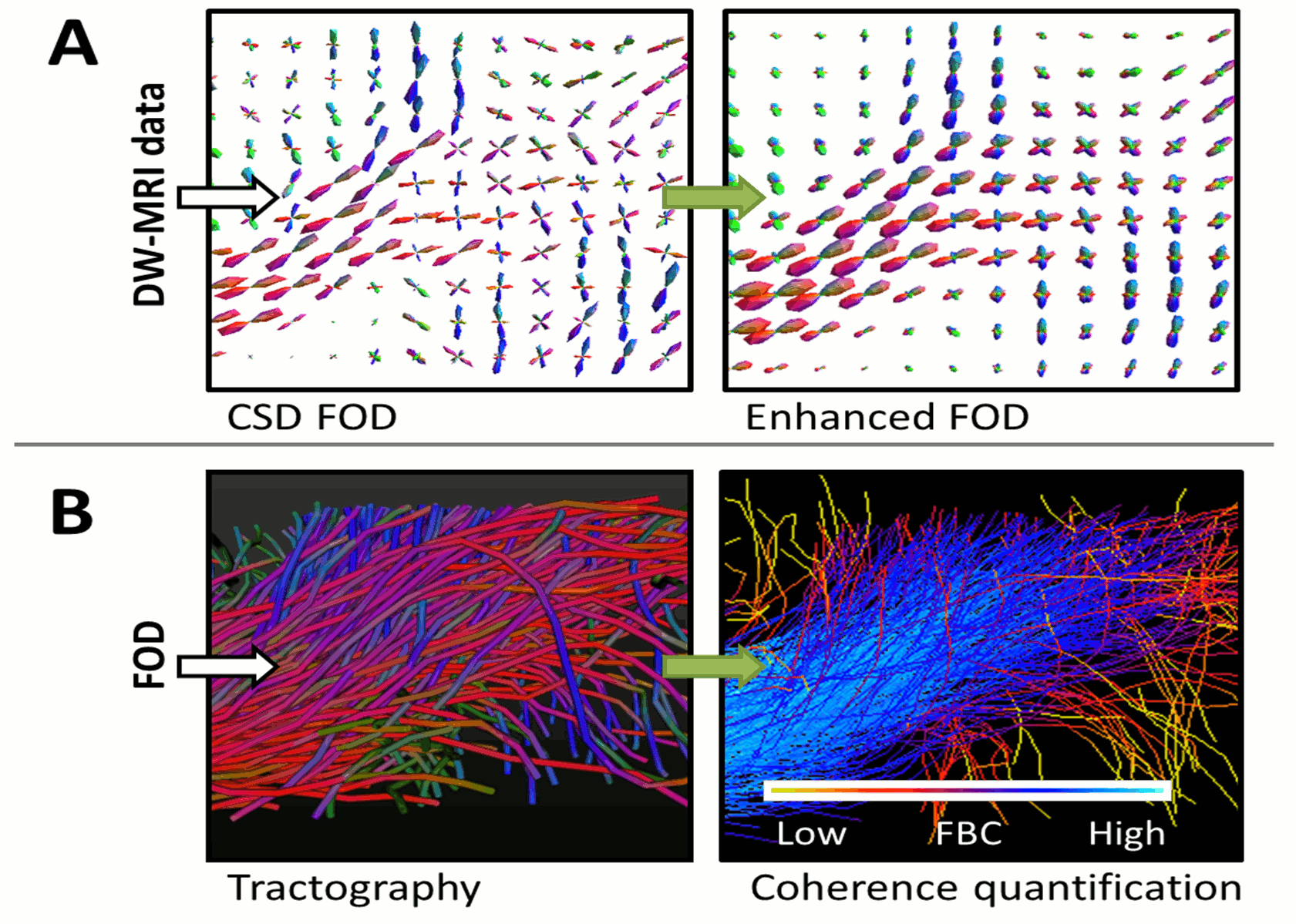}
\caption{{\bf The proposed pipeline of the paper.} 
 CSD is used to estimate an FOD from DW-MRI data. The FOD is enhanced (\textbf{A}) with PDE techniques. Then a deterministic or probabilistic tractography is applied to the (enhanced) FOD (probabilistic shown here, with coloring indicating the fiber direction). In the lower right figure, we applied our coherence quantification method (\textbf{B}), based on the same PDE framework, which shows that blue fibers are well aligned (high Fiber to Bundle Coherence (FBC)) and yellow fibers are spurious (low FBC). The green arrows indicate the steps in which the contextual PDEs are used.
 }\label{fig:csdenhpipeline}
\end{figure}

Tractography methods are often used in the DW-MRI pipeline to provide insight in the structural connectivity of the white matter bundles. Independently of the model used for interpreting the DW-MRI data, noise originating from the scanner, acquisition artifacts and partial volume effects \cite{Jones2010} are likely to result in spurious (aberrant) fibers in the tractography output. To improve the data on which the tractography is performed, different regularization methods can be used. Methods exist that apply filtering for the reduction of noise directly on the DW-MRI data \cite{Wiest2007,Coupe2008,Descoteaux2008}, other methods aim to regularize the DTI tensor fields \cite{Poupon2000,Coulon2001,tschumperle2001,Burgeth2009a,Burgeth2009b}. On HARDI data the regularization can be performed on individual voxels \cite{Tuch2004,Descoteaux2007,descoteaux2009deterministic} or in combination with the local spatial information \cite{Barmpoutis2008,Goh2009,schultz2012,Reisert2011,Tax2014}.

\begin{figure}[t!]
  \centering
  \includegraphics[width=0.95\textwidth]{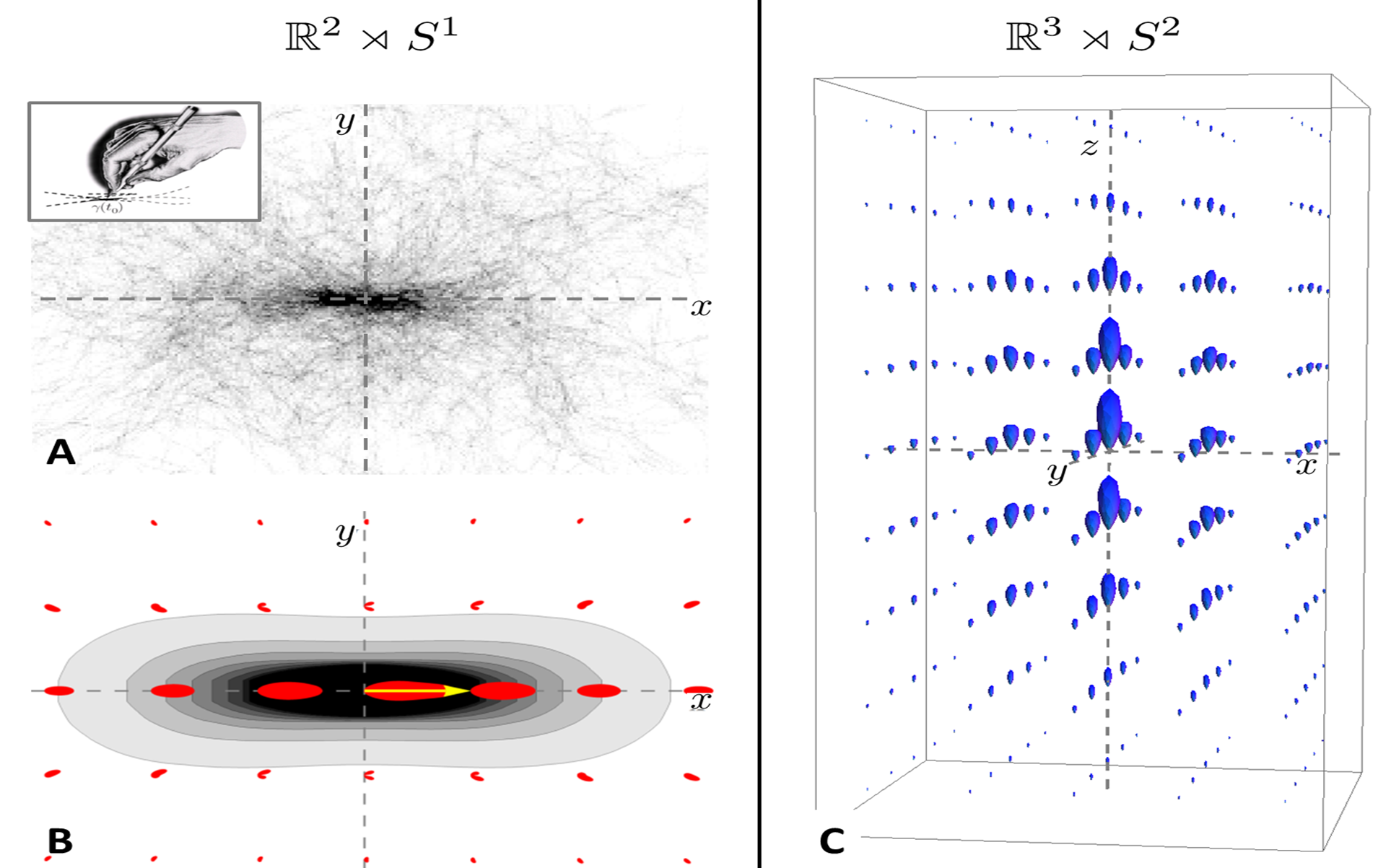}
\caption{{\bf Stochastic interpretation of the contour enhancement kernels.}
\textbf{A.} Accumulation of 300 sample paths drawn from the underlying stochastic process of the contour enhancement PDE in $\mathbb{R}^2 \rtimes S^1$, projected on the xy-plane. \textbf{B.} The contour enhancement kernel arises from the accumulation of infinitely many sample paths. The gray-scale contours indicate the marginal of the kernel, obtained by integration over $S^1$, the red glyphs are polar graphs representing the kernel at each grid point. \textbf{C.} The contour enhancement kernel oriented in the positive $z$-direction in $\mathbb{R}^3 \rtimes S^2$ can be visualized on a grid with glyphs that in this case are spherical graphs.}\label{fig:kernelvis}
\end{figure}

We introduce two new strategies based on the same underlying principle to improve fiber alignment in tractography results, in order to have more reliable information on the structural connectivity of brain. First we perform contextual regularization to the FOD obtained with CSD, see Fig. \ref{fig:csdenhpipeline}A, and secondly we introduce a fiber to bundle coherence (FBC) measure that can be applied to any fiber bundle to classify and remove spurious fibers, see Fig. \ref{fig:csdenhpipeline}B. Both approaches are based on a partial differential equation (PDE) framework introduced in \cite{Franken2008,Duits2011,Creusen2011,Duits2013}, where the Fokker-Planck equation of a stochastic process for enhancement of elongated structures is considered. These type of PDE-based enhancement methods have been widely used for the processing of 2D-images. In this framework, images are represented in the extended space of positions and orientations via a stable invertible orientation score \cite{franken2009crossing}, that associates to every location an orientation distribution of the local image features (lines and contours). Then, the stochastic processes for contour completion \cite{Mumford1994,Zweck00euclideangroup,SanguinettiJOV2010,August2003,Duits2008ESL,Momayyez2013} and contour enhancement \cite{Citti2006,Duits2010LIP1,Agrachev2009} (see Fig. \ref{fig:kernelvis}A) on this extended space $\mathbb{R}^2 \rtimes S^1$ induce crossing preserving completion and enhancement of lines \cite{franken2009crossing}. 

The DW-MRI data that we use is naturally defined on the coupled space $\mathbb{R}^3 \rtimes S^2$ of 3D positions and orientations. As in the 2D case, crossing preserving enhancement of line structures is required, for which we use the 3D extension of the 2D stochastic process for contour enhancement, introduced in \cite{Duits2011}. The linear PDE corresponding to this stochastic process can be solved by convolution of the initial condition with the kernel of the PDE. This kernel is also a function on the position-orientation space and can be seen as a transition distribution from the origin (in position and orientation) to neighboring elements. From the stochastic point of view, the kernels arise as limits of the accumulation of infinitely many sample paths drawn from the stochastic process, illustrated in Fig. \ref{fig:kernelvis}A. For mathematical details of the underlying stochastic processes of the PDEs, see \cite[\textsection 10.1]{Duits2013}. The general idea needed for this article is sketched in Fig. \ref{fig:kernelvis}. In Figs. \ref{fig:kernelvis}B and \ref{fig:kernelvis}C we show the contour enhancement kernel using glyph visualization on a grid, each glyph being a polar (red, 2D) or spherical (blue, 3D) graph plot where in every orientation the (spherical) radius is proportional to the value of the kernel. This type of visualization is used throughout the paper for functions defined on the space of positions and orientations.

Recently, many authors \cite{Momayyez2013,Vesna2010,Vesna2015,Reisert2012,Reisert2013,Tax2014,Duits2011,Duits2013,DelaHaije2014} demonstrated the advantages of contextual processing of DW-MRI data. The general rationale behind contextual processing is to include alignment of local orientations and their surroundings (i.e. the context) on the coupled space of positions and orientations. For this alignment of local orientations, roto-translations are needed, which imposes a non-Euclidean structure in the PDE-based processing as we explain in Section \ref{se:enhancement}. More details on the embedding of $\quotpo$ in the roto-translation group $SE(3)$ can be found in \cite{Duits2011}. This demonstrates how either the completion or enhancement PDEs can be used to extrapolate DTI information to increase the angular resolution and resolve some fiber crossings. This idea was shown to be promising in clinical experiments \cite{Vesna2010,Vesna2015}, but in some cases extreme parameters had to be set to obtain clear maxima at crossings (where DTI data is inadequate). Therefore in this paper we introduce and test the combination of CSD with contextual enhancements. The method proposed in \cite{Momayyez2013} uses an advection-diffusion equation (that we called contour completion above) to improve HARDI data to obtain connectivity measures. In our work we rely on a purely diffusive process, contour enhancement, which in contrast to contour completion does not suffer from singularities \cite{Duits2013} and is less sensitive to small perturbations of the initial conditions. This property makes the enhancement process more suited to be combined with the sharp angular distributions produced by CSD. As the methods mentioned above still result in broad angular distributions, they need to be combined with some sharpening method. To this end, a geometric morphological sharpening based on erosions was presented in \cite{Duits2013,DelaHaije2014,Tax2014}. 
Another related method presented in \cite{Reisert2012,Reisert2013} is the so-called fiber continuity model in which purely spatial regularization is considered in combination with spherical deconvolution as alternative to the non-negativity constraint in the classical CSD \cite{Tournier2008}. In Section \ref{se:enhancement} we demonstrate the importance of including also an angular regularization term.

\subsection{Contributions}
The first contribution of this article is to study the combination of the widely used CSD method with a regularization induced by the enhancement PDE acting on the FOD. Since the FOD obtained with CSD consists of sharp angular profiles, it is well-suited as an initial condition for the enhancement PDE, that typically has a smoothing effect on the orientation distributions. The contextual regularization method reduces non-aligned crossings in the FOD, allowing for a better alignment of fibers when tracking is applied on the enhanced FOD. We show that this method is therefore useful to reduce the number of false positive fibers, but mainly to find more true positives in the tractography output. Although in this paper we compare to the classical CSD method, the PDE enhancements can also be applied to extensions of this method \cite{Tax2013,schultz2013auto,Jeurissen2014411,10.3389/fninf.2014.00028,Roine2015}.

The second contribution of this article is to introduce the fiber to bundle coherence (FBC) measure. The motivation for this measure is that, especially probabilistic, tracking methods typically produce spurious fibers that should be removed from the tractography. In contrast to the first approach, this method serves as a post-processing tool. For the computation of the FBC we regard the fiber bundle as a set of oriented points, by considering for every fiber point also the local tangent to the fiber. We construct a density using the enhancement PDE with an initial condition that is a sum of superposed $\delta$-distributions at every oriented point in the bundle. The construction of such a density from tracks relates to track density imaging \cite{calamante2010track} and track orientation density imaging \cite{dhollander2014track}, though here the use of the contour enhancement kernels, Fig. \ref{fig:kernelvis}, allows to use a sparse set of fiber tracks. The FBC, a measure for spuriousness of fibers, is computed by efficient integration of this fiber-based density. Fibers that are most spurious according to the FBC can be removed from the tractography, resulting in a better aligned fiber bundle. Complementary to the first method, this FBC measure has the purpose to remove false positives in a tractography.

\subsection{Structure of the Article}
Section \ref{se:methods} covers theory of the individual parts of the pipeline as outlined in Fig. \ref{fig:csdenhpipeline}, consisting of CSD, PDE enhancements, tractography and coherence quantification in Sections \ref{se:CSD}-\ref{se:coherence}, respectively. In Section \ref{se:results} we provide extensive validation of the combination of CSD and PDE enhancements and the FBC, using three experiments: 
\begin{enumerate}
	\item First we use the Tractometer evaluation system \cite{cote2012tractometer,Cote2013} on the ISBI 2013 HARDI reconstruction challenge dataset \cite{Daducci2014}, a digital phantom with known ground truth, to demonstrate how contour enhancement improves both the local FOD reconstruction and the global connectivity of fiber bundles compared to CSD, see Section \ref{se:HARDIrecon}. 
  	\item In Section \ref{se:evaluationDWMRI} we show on a human DW-MRI dataset, containing different crossing bundles, that CSD combined with enhancements yields an FOD that is more robust with respect to the $b$-value and the number of gradient directions used in the acquisition. Furthermore, we make a comparison with earlier work involving erosions and nonlinear diffusion of FODs directly applied to a DTI-model \cite{Duits2013,Tax2014}, that was based on the same data. We show that with our method the glyphs are sharper at the locations where bundles cross.
  	\item Finally in Section \ref{se:ORexperiment}, we show an experiment with clinical data in which we reconstruct the optic radiation (OR) to determine the position of the tip of the Meyer's loop, that is of interest in epilepsy surgery planning \cite{falconer1963follow,powell2005mr,Sherbondy2008Contrack,Tax2014,Meesters2013}. Accurate estimation of this position is difficult due to the presence of spurious fibers in the reconstruction of the OR. We show that both the FOD enhancement and the FBC measure, see Fig. \ref{fig:csdenhpipeline}, and in particular the combination of the two allow for a more stable determination of the tip of the Meyer's loop. Here `more stable' means less variation with respect to stochastic realizations in the probabilistic tractography results.
\end{enumerate}
Conclusions and a discussion can be found in Section \ref{se:conclusions}.

\section{Methods}\label{se:methods}
In this paper it is assumed that we have HARDI data as input, from which we derive an FOD $U$ that models the orientation of fibers in each voxel, i.e. $U: \mathbb{R}^3 \times S^2 \rightarrow \mathbb{R}^+$. For this we use CSD \cite{Tournier2012}, concisely described in Section \ref{se:CSD}, as it gives sharp angular profiles and is able to distinguish multiple fiber directions within a voxel.

Then we use the enhancement PDE for diffusion of the FOD $U$, coupling spatial and angular information. The combination of CSD and such enhancement is a powerful method to obtain an enhanced FOD in which the coherence inherent in the data is included, providing a more coherent input for the tractography. The enhancement technique is explained in Section \ref{se:enhancement}.

We use the MRtrix algorithm \cite{Tournier2012} for both deterministic and probabilistic tractography to estimate the structural connectivity in the brain. In the deterministic tractography, fiber tracks are obtained by integrating a directional field, given an initial position and direction. The directional field is given by the locally maximal orientations in the glyphs. In contrast to deterministic tractography, the probabilistic tractography method of MRtrix samples the orientations from the entire FOD and does not use just the maxima. More difficult paths can be reconstructed than with deterministic tracking, but typically also many spurious fibers are produced due to the probabilistic sampling. Both the deterministic and the probabilistic method are explained in more detail in Section \ref{se:tractography}.

In Section \ref{se:coherence} we introduce our new technique to quantify the coherence of fibers with respect to all the fibers in a bundle, based on the same PDE theory as employed for the contextual enhancement in Section \ref{se:enhancement}. We explain how the kernel of the enhancement PDE is used to construct a tractography-based density, how the FBC is computed and how this measure is able to classify spurious fibers in a tractography.

\subsection{A Brief Review of CSD}\label{se:CSD}

In CSD it is assumed that at each voxel position $\by$ the measured signal $S_{\by}:S^2 \rightarrow \mathbb{R}$ can be represented by a spherical convolution of the FOD $f_{\by}:S^2 \rightarrow \mathbb{R}$ with a response function $K:S^2 \rightarrow \mathbb{R}$, that is estimated from the data \cite{Tournier2004}. Since the spherical deconvolution to determine the FOD is ill-posed, a non-negativity constraint is included as in \cite{Tournier2007,Tournier2008}. Then, given the signal $S_{\by}(\bn)$ for a sample of orientations $\bn \in S^2$, the solution of CSD is found by iteratively solving the minimization problem:

\begin{equation}\label{eq:contCSD}
 f^{i+1}_{\by}(\bn)=\underset{g \in \mathbb{L}_2(S^2)}{\textmd{argmin}}\underbrace{\int_{S^{2}}|(K\ast_{S^{2}}g)(\textbf{n})-S_{\by}(\textbf{n})|^{2}d\sigma(\textbf{n})}_\text{Data Driven}+ \lambda^{2} \underbrace{\int_{S^{2}}|(\mathcal{L}_{f_{\by}^{i}}(g))(\textbf{n})|^{2}d\sigma(\textbf{n})}_\text{Regularization},
\end{equation}
for $i = 1, \dots, i_{\text{max}}$, with $i_{\text{max}}$ the maximum number of iterations. Here $K \in \mathbb{L}_2(S^2)$ is aligned with and symmetric around the $z$-axis, the convolution $\ast_{S^2}$ is the usual $S^2$ spherical convolution \cite{Driscoll1994S2conv}, $d \sigma(\textbf{n})$ is the Jacobian of the surface measure in orientation $\textbf{n}$ and $\lambda$ is a parameter to influence the trade-off between the data driven term and regularization term. The linear operator $\mL_h: \mathbb{L}_2(S^2) \rightarrow \mathbb{L}_2(S^2)$ in the regularization term gives the non-negativity constraint and is defined by:

\begin{equation}
  (\mL_h f)(\bn) = f(\bn)H(\tau_h - h(\bn)), \qquad \text{for given } h \in \mathbb{L}_2(S^2),
\end{equation}
where $H$ is the Heaviside function and $\tau_h$ is a threshold equal to a fixed factor $\tau$ times the mean of $h$. The initial function $f_{\by}^0$ for the iteration is computed by taking only the data driven term of Eq. (\ref{eq:contCSD}). The iteration stops when successive iterations yield the same result, typically after 5 to 10 iterations \cite{Tournier2007}. Throughout the paper, we call $U$ the FOD obtained by

\begin{equation}
  U(\by,\bn) = f^{i_{\text{max}}}_{\by}(\bn).
\end{equation}

In practice CSD is performed using spherical harmonics with a maximal spherical harmonic order of $8$ ($l_{\textmd{max}}=8$) as discussed in \cite{Tournier2013}.

Improvements to the original CSD exist to modify and improve the response function, either by recursive calibration or auto-calibration \cite{Tax2013,schultz2013auto}, by using multiple acquisition shells \cite{Jeurissen2014411} or by including anatomical data \cite{10.3389/fninf.2014.00028,Roine2015}. The latter two methods aim to reduce the partial volume effects, where CSD is likely to produce spurious fiber orientations. These partial volume effects can occur when in a voxel multiple tissues or multiple bundles with different orientation are present. Here we use the classical CSD as it is the basic technique available in several neuroimaging packages. However, we stress that our method is not restricted to this type of CSD. In any case, our method aims to reduce non-aligned crossings in the FOD, also the ones induced by partial volume effects, as we will show in several experiments in this paper. Further improvement of the methodology can be expected when including recently extended and more elaborate CSD techniques \cite{schultz2013auto,Jeurissen2014411,Roine2015}, but this is left for future work.

\subsection{Contour Enhancement (Step A)}\label{se:enhancement}
\begin{figure}[t!]
\centering
  \includegraphics[width=0.6\textwidth]{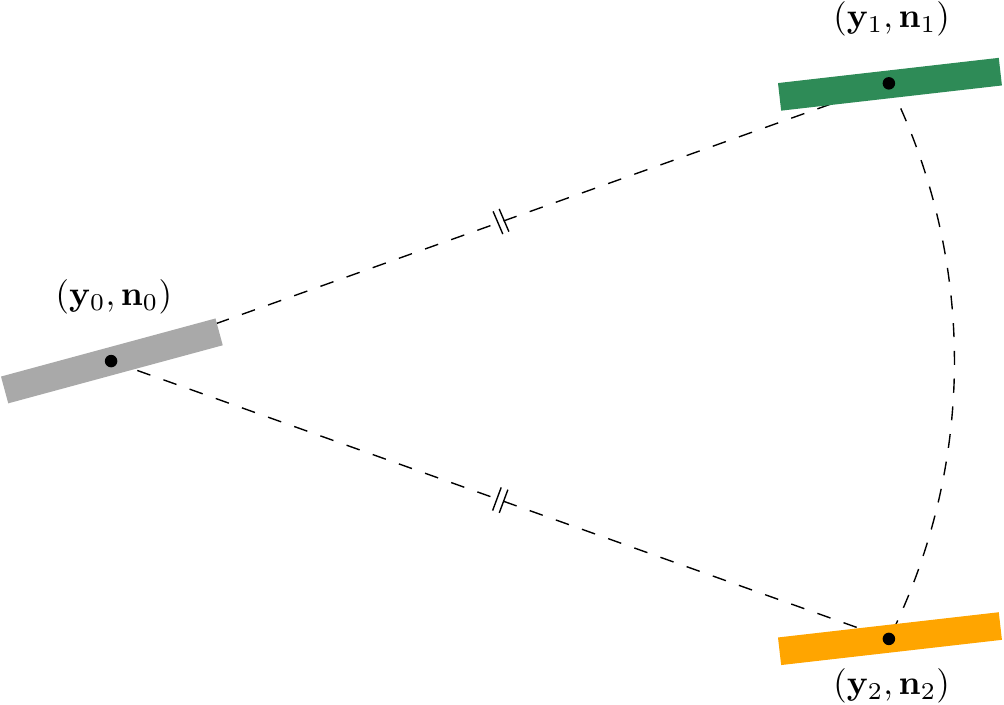}\caption{{\bf The concept of alignment requires a coupling of positions and orientations.}
  The pair of position and orientation $(\by_0,\bn_0)$ is better aligned with $(\by_1,\bn_1)$ than with $(\by_2,\bn_2)$, even though spatial and angular distances are equal. Formally we can say that the sub-Riemannian distance on $\mathbb{R}^3\rtimes S^2$ \cite{Duits2013} is smaller between $(\by_0,\bn_0)$ and $(\by_1,\bn_1)$.}\label{fig:coupling}
\end{figure}

To improve alignment of neighboring glyphs of the FOD $U$, recall the glyph field visualization in Figs. \ref{fig:csdenhpipeline} and \ref{fig:kernelvis}C, we apply contextual enhancements. Before we specify the PDE we consider for this enhancement, we first need to express the notion of alignment in mathematical terms. To this end, let us consider Fig. \ref{fig:coupling}, where it is shown that the notion of alignment cannot be supported by a decoupled, flat Cartesian product $\mathbb{R}^3 \times S^2$ with the combined Euclidean distance. It is clear that the green bar at $(\by_1,\bn_1)$ is better aligned with the gray bar at $(\by_0,\bn_0)$ than the orange bar at $(\by_2,\bn_2)$, even though the distances in the space $\mathbb{R}^3 \times S^2$ are equal, i.e. $\sqrt{d_{\mathbb{R}^3}^2(\by_0,\by_1)+d_{\mathbb{S}^2}^2(\bn_0,\bn_1)}=\sqrt{d_{\mathbb{R}^3}^2(\by_0,\by_2)+d_{\mathbb{S}^2}^2(\bn_0,\bn_2)}$. This means that in order to appropriately describe the concept of alignment, we must consider more than just the amount of spatial displacement and the amount of change in orientation. Coupling these two types of motion (via rigid body motions) is a solution to this problem \cite{Duits2011}. The coupling follows very naturally by expressing the motion of an oriented particle $(\by,\bn)$ in terms of a moving frame of reference determined by its orientation. That is, spatial movement along the orientation $\bn$ should be much cheaper than spatial movement in the plane orthogonal to $\bn$. This creates a natural anisotropy for spatial movement. For angular motion we need isotropy. This extra structure can be obtained by embedding the space of positions and orientations in the rigid body motion group. This means that an element $(\by,\bn) \in \mathbb{R}^3 \times S^2$ is identified with the rigid body motion $(\by,\bR_{\bn})$, where $\bR_{\bn}$ is \emph{any} rotation matrix such that $\bR_{\bn}\be_z = \bn$, with $\be_z \in S^2$ pointing to the north pole. We denote this space of coupled positions and orientations by $\mathbb{R}^3 \rtimes S^2$, so we have
\begin{equation}
  \mathbb{R}^3 \times S^2 \ni (\by, \bn) \leftrightarrow (\by,\bR_{\bn}) \in \mathbb{R}^3 \rtimes S^2.
\end{equation}
 The group $\mathbb{R}^3 \rtimes S^2$ is equipped with the following (non-commutative) group product:

\begin{equation}\label{eq:groupproduct}
  (\by, \bR)(\by', \bR') = (\by + \bR \by', \bR \bR').
\end{equation}
This product moves oriented elements in a shift-twist fashion, rather than by a rotation followed by an independent translation. Due to this shift-twist group product in Eq. (\ref{eq:groupproduct}), we automatically express motion of oriented particles in terms of a moving frame in $\mathbb{R}^3 \rtimes S^2$, which makes this space well-suited for the application of our contextual enhancements. Nevertheless, in the remainder of this article this space can still be regarded as the Cartesian 5D space $\mathbb{R}^3 \times S^2$, where we secured the coupling of positions and orientations via our specific choice of differential operators and diffusions that are applied.

\begin{figure}[t!]
  \centering
  \includegraphics[width=1.0\textwidth]{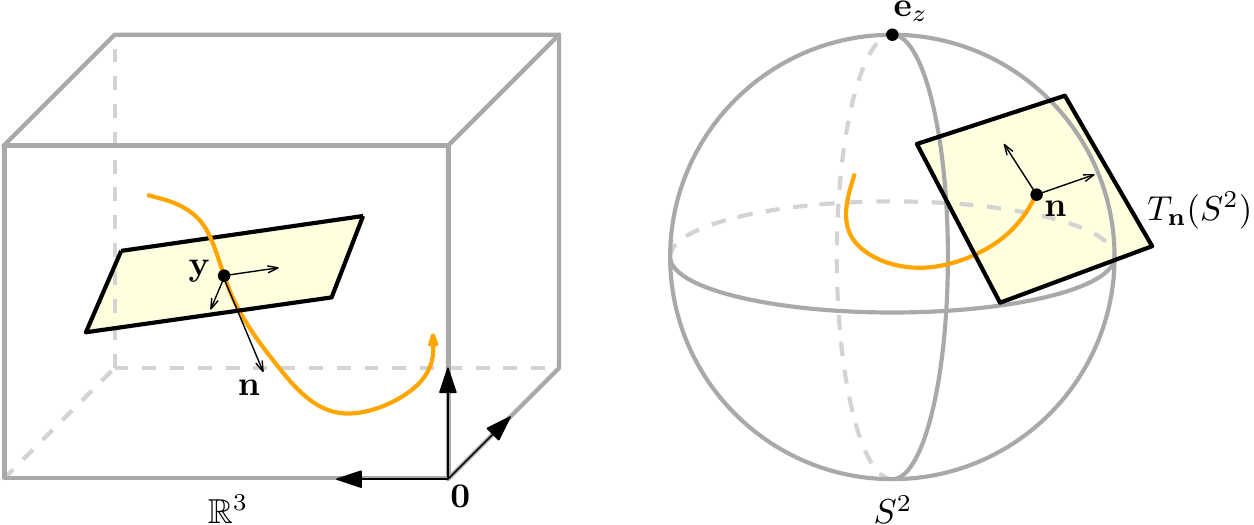}
\caption{{\bf Diffusion in the space of positions and orientations.}
Spatial diffusion is applied in the direction of the fiber (left), angular diffusion is applied in the tangent plane perpendicular to the fiber direction (right).}\label{fig:r3s2diff}
\end{figure}

To improve alignment of FOD glyphs, we use a particular diffusion process called contour enhancement that uses both spatial and angular diffusion in the extended space of positions and orientations \cite{Duits2011}. Given a structure (think of a fiber bundle) in this space, see Fig. \ref{fig:r3s2diff}, we apply spatial diffusion only in the direction of the structure, not in the spatial plane perpendicular to it. Angular diffusion is applied in the plane tangent to $S^2$ at the point $\bn$. This diffusion process enhances elongated structures, while preserving crossing structures, and is given by a Fokker-Planck type of system, a linear diffusion equation on $\mathbb{R}^3 \rtimes S^2$. For $t \geq 0$, $\by \in \mathbb{R}^3$, $\bn \in S^2$ this system can be expressed as:
\begin{equation}\label{eq:contenh}
\left\{
\begin{aligned}
    &\partial_t W(\by,\bn,t) = \left(D_{33}(\bn \cdot \nabla_{\by})^2 + D_{44} \Delta_{S^2} \right)W(\by,\bn,t), \\
    & W(\by,\bn,0) = U(\by,\bn).
  \end{aligned} \right.
\end{equation}
Here $W(\by,\bn,t)$ is a scale space representation in $(\mathbb{R}^3 \rtimes S^2)\times\mathbb{R}^+$ \cite{Duits2007}. The symbol $\nabla_{\by}$ denotes the gradient with respect to the spatial variables and $\Delta_{S^2}$ is the Laplace-Beltrami operator on the sphere. Parameters $D_{33}>0$ and $D_{44}>0$ are related to the amount of spatial and angular diffusion, respectivly. Parameter $t\geq 0$ is the diffusion time of the contour enhancement process. It can be seen as a Brownian motion process, recall Fig. \ref{fig:kernelvis}A, where particles are allowed to spatially move back and forth in the direction they are heading, or change their direction, but are not allowed to step aside (comparable to the movement of a car).

 We refer to the solution of Eq. (\ref{eq:contenh}) as the enhanced FOD. It can be obtained via a finite difference scheme \cite{Creusen2013}, or via a convolution with a kernel $p_t:\mathbb{R}^3 \rtimes S^2 \rightarrow \mathbb{R}^+$:

\begin{equation}\label{eq:shifttwistconv}
\begin{aligned}
  W(\by,\bn,t) &= \left(p_t \ast_{\mathbb{R}^3 \rtimes S^2} U  \right)(\by,\bn)\\
  &= \int_{S^2} \int_{\mathbb{R}^3} p_t ((\bR_{\bn'})^T(\by - \by'), \bR_{\bn'}^T \bn) \cdot U(\by', \bn') \; \rmd \by' \rmd \sigma(\bn').
  \end{aligned}
\end{equation}
A basic approximation to the exact Green's function of the contour enhancement PDE is known \cite{Duits2011} and can be written as the product of Green's functions $p_t^{\mathbb{R}^2 \rtimes S^1}$ in the following way:

\begin{equation}\label{eq:kernR3S2}
\begin{aligned}
  p_t(\by, \bn) = \frac{8}{\sqrt{2}} D_{33}t \sqrt{\pi t D_{44}} \cdot & p_t^{\mathbb{R}^2 \rtimes S^1}(z/2,x,\beta) \cdot  p_t^{\mathbb{R}^2 \rtimes S^1}(z/2,-y,\gamma),
  \end{aligned}
\end{equation}
with $\bn = \bn(\beta,\gamma) = \bR_{\be_x,\gamma} \bR_{\be_y,\beta} \be_z = (\sin \beta, -\cos \beta \sin \gamma, \cos \beta \cos \gamma)^T$, $\beta \in [-\pi, \pi)$, $\gamma \in (-\frac{\pi}{2}, \frac{\pi}{2})$. The $\mathbb{R}^2 \rtimes S^1$ kernels are given by

  \begin{equation}
    p_t^{\mathbb{R}^2 \rtimes S^1}(x,y,\theta) = \frac{1}{32 \pi t^2 D_{44}D_{33}}e^{-\sqrt{\frac{EN(x,y,\theta)}{4t}}},
  \end{equation}
  with 

  \begin{equation}
    EN(x,y,\theta) = \left(\frac{\theta^2}{D_{44}} + \frac{1}{D_{33}}\left(\frac{\theta y}{2} + \frac{\theta/2}{\tan(\theta/2)}x\right)^2 \right)^2 + \frac{1}{D_{44}D_{33}} \left(\frac{-x \theta}{2}+\frac{\theta/2}{\tan(\theta/2)}y\right)^2.
  \end{equation}
  To avoid numerical errors, we use the estimate $\frac{\theta/2}{\tan(\theta/2)} \approx \frac{\cos(\theta/2)}{1- (\theta^2/24)}$ for $|\theta|<\frac{\pi}{10}$. This approximation is easy to use and allows for efficient implementation \cite{rodrigues2010}. 

\begin{figure}[t!]
        \centering
        \includegraphics[width=\textwidth]{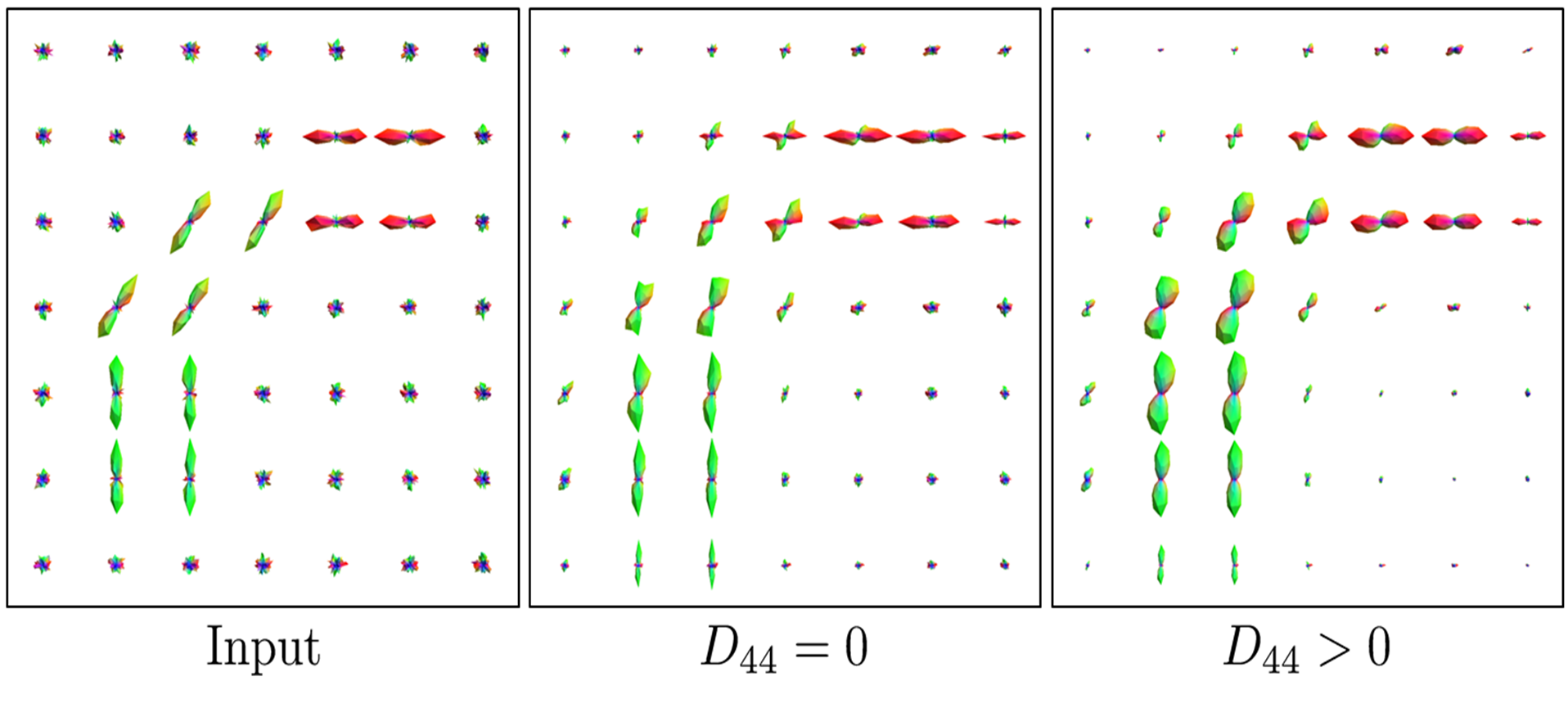}
        \caption{{\bf The importance of including angular diffusion.}
        (Left) artificial input data of a curved bundle. (Middle) diffusion with $D_{33}>0$, $D_{44}=0$. (Right) contour enhancement with $D_{33},D_{44}>0$. Fiber propagation with $D_{44}=0$ leads to crossing artefacts rather than smooth fiber enhancement.}\label{fig:motivationD44}
\end{figure}

From the approximation kernel in Eq. (\ref{eq:kernR3S2}) it can be seen that problems could occur when either $D_{33}=0$ or $D_{44}=0$. To this end, a necessary and sufficient condition for the existence of a smooth solution kernel for the evolution process in Eq. (\ref{eq:contenh}) is given by the H\"ormander requirement \cite{Hormander1967}. This condition applies to more general situations than the one here, see e.g. \cite{Duits2011}, but for the specific case of contour enhancement the requirement is satisfied iff $D_{33},D_{44} > 0$. Setting $D_{44}=0$ would result in a singular non-smooth kernel, which has numerical disadvantages. More importantly, apart from this theoretical issue the need for both spatial and angular diffusion can also be argued from a practical point of view, as is illustrated in Fig. \ref{fig:motivationD44}. We use an artificial example in which a curved fiber bundle is present, shown in the left figure. When the input is diffused with $D_{44}=0$ as in the middle of Fig. \ref{fig:motivationD44}, the peaks stay distinct and point in the wrong direction. On the other hand, when $D_{44}>0$ as in the right figure, due to the angular diffusion the peak is redirected and the glyphs lie better aligned with the fiber bundle. Hence $D_{44}>0$ is needed to ensure the crucial interaction between different orientations. Finally we recall the relation between Tikhonov regularization and diffusion, see e.g. \cite[Thm 2]{Duits2011}, which allows us to connect diffusion with $D_{44}=0$ with the fiber continuity model in \cite{Reisert2012,Reisert2013}. This model does not suffer from the inconvenience of considering only spatial regularization, as they represent the FOD in a truncated spherical harmonic basis. When the enhancements are used in combination with probabilistic tractography, we first apply a standard sharpening deconvolution transform to the FOD as described in \cite{descoteaux2009deterministic}, to maintain the sharpness of the FOD.

\subsection{Tractography}\label{se:tractography}
As the next step in the pipeline we use the MRtrix tractography algorithm \cite{Tournier2012}, as implemented in http://www.brain.org.au/software/index.html\#mrtrix, version 0.2.12. It allows us to perform deterministic and probabilistic fiber tracking on spherical harmonic representations of the (enhanced) FOD. To have a fair comparison between trackings on the FOD and the enhanced FOD, we use the parameter settings as explained next. 

\begin{itemize}
\item  In the deterministic tracking of MRtrix, seed points are randomly selected from a seed region. The initial direction is sampled randomly and every next step follows the direction of the most aligned FOD maximum. If this maximum is below a threshold value, the fiber terminates. This threshold (cutoff) is set to 10\% of the maximal angular response of the FOD. There is no constraint on the maximal curvature of the fibers. To prevent that fibers have an initial direction that is not aligned with the fiber bundle, we force the initial direction to be approximately in the direction of the maximal FOD peak, by setting the initial cutoff to 0.9. The step size is set to $1/10$th of the voxel size as is suggested in \cite{Tournier2012}. Tracks proceed in both directions from the seed point and terminate either when they hit the boundary of the volume or mask (if applicable), or due to the threshold stopping criterion. 
\item In the probabilistic case, starting from the seed region, every next step follows a direction randomly sampled from the FOD. Here we set the minimal radius of curvature to $1$ mm, the default value in the MRtrix algorithm. Optionally, a target region of interest is used to select only those fibers that cross this region.
\end{itemize}

We base our choice for deterministic tractography or probabilistic tractography on the application. If only a seed region is specified, as in Sections \ref{se:HARDIrecon} and \ref{se:evaluationDWMRI}, we use deterministic tractography. In this case there is too much freedom in the probabilistic algorithm and the streamlines show a lot of spurious behavior. Here, a probabilistic approach could make sense if extreme amounts of tracks are used for track density methods. As we do not pursue these methods here, we prefer to use deterministic tractography. If both a seed region and an end region are specified, as in the optic radiation application in Section \ref{se:ORexperiment}, we prefer to use probabilistic tractography. It is known that deterministic tractography in this case provides only a few of the possible pathways from the seed to the end region, whereas reconstructions with probabilistic tractography are much fuller. 

Probabilistic tractography results typically contain many false positive fibers. Streamlines that are anatomically implausible can be removed with scoring methods \cite{Tax2014,Sherbondy2008Contrack} or by imposing anatomical constraints. Even when using these methods, the filtered tractography output can still contain fibers that deviate from the fiber bundle and are likely to be spurious. In the next section, we propose a coherence measure for fibers in a fiber bundle in order to classify these spurious fibers.

\subsection{Coherence Quantification of Fiber Bundles (Step B)}\label{se:coherence}
\begin{figure}[t!]
  \centering
  \includegraphics[width = \textwidth]{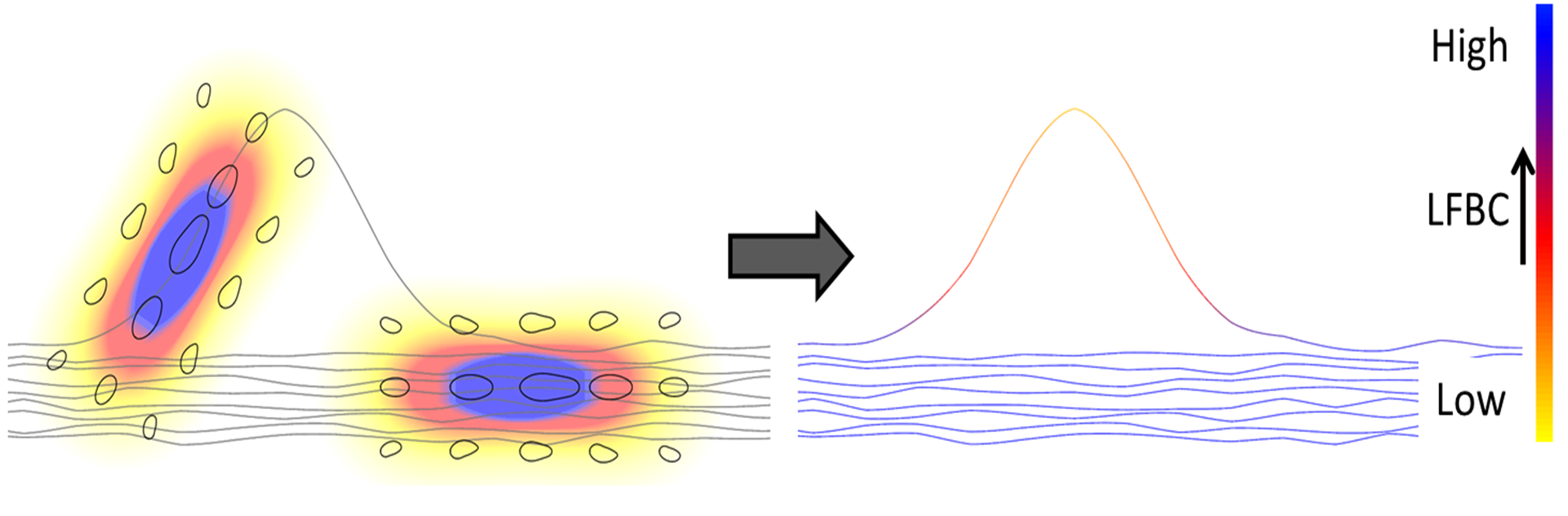}
\caption{{\bf Construction of the LFBC.}
The local fiber to bundle coherence (LFBC) is constructed for a set of fibers (gray lines), illustrated in 2D for simplicity, as follows. Every local tangent in the tractography contributes to the density, by considering it as a $\delta$-distribution. We convolve this with the contour enhancement kernel, shown on the left for two points, visualized as in Fig. \ref{fig:kernelvis} and with the coloring indicating the contribution of the kernel to the LFBC. Doing so for all points, fiber points that are isolated from or badly aligned with other fibers receive low contributions, such as the outlying fiber. The LFBC along the fibers is displayed on the right.}\label{fig:spurfibschematic}
\end{figure}

 In this section we introduce our second contribution of the paper, a \emph{fiber to bundle coherence} (FBC) measure to quantify the coherence of each fiber with respect to all other fibers in the bundle, recall Fig. \ref{fig:csdenhpipeline}B. A spurious fiber, as schematically shown in Fig. \ref{fig:spurfibschematic}, is isolated from or poorly aligned with the bulk of the tracks and is therefore unlikely to represent the underlying brain structure.  Fibers with low coherence, i.e. a low FBC, can then be classified as spurious.

To classify a fiber as spurious, we first construct a density by regarding each fiber as a superposition of $\delta$-distributions in $\mathbb{R}^3 \rtimes S^2$ and convolving this distribution with the kernel in Eq. (\ref{eq:kernR3S2}). This density is independent of the underlying data and is based purely on the collection of fibers $\Gamma$. Integration of this density along a part of length $\alpha$ of a fiber gives a local measure for the coherence of that part.

Next we explain the mathematical techniques that support the idea in Fig. \ref{fig:spurfibschematic}. We denote the fibers from a tractography output by $\by_i(s) \in \mathbb{R}^3$, $1 \leq i \leq N$, $0 \leq s \leq l_i$, with $s$ the arc length parameter, $l_i$ the total length of fiber $i$ and $N$ the number of fibers. Now let $\bn_i(s) = \dot{\by}_i(s)$ be the tangent of the fiber, so that $\gamma_i(s) = (\by_i(s), \bn_i(s))$ forms a curve (fiber) in $\mathbb{R}^3 \times S^2$. By construction, $\bn_i(s)$ points in the forward direction of the fiber. Since in DW-MRI data antipodal orientations are identified, we also consider $\bar{\gamma}_i(s)= (\by_i(s),-\bn_i(s))$. The complete fiber bundle is defined as $\Gamma := \{\gamma_i \;|\; i = 1,\dots,N\} \cup \{\bar{\gamma}_i \;|\; i = 1,\dots,N\}$.
A discrete formulation of a fiber $i$ with $N_i$ points is given by:

\begin{equation}\label{eq:notationgamma}
\gamma_i^j := \gamma_i(s^j) = (\by_i(s^j), \bn_i(s^j)) =: (\by_i^j,\bn_i^j), \quad s^j = \frac{j-1}{N_i-1}l_i, \quad j = 1,\dots,N_i.
\end{equation}

 This way there are $N_{tot} = 2 \sum_{i=1}^N N_i$ elements in $\Gamma$. Now we regard every point $\gamma_i^j$ as a $\delta$-distribution in $\mathbb{R}^3 \rtimes S^2$ centered around $(\by_i^j,\bn_i^j)$. A density for the entire bundle is then constructed as follows:
\begin{equation}
  F_{\Gamma}(\by,\bn) = \frac{1}{N_{tot}} \sum_{\sigma=1}^2 \sum_{i=1}^N \sum_{j=1}^{N_i}  \delta_{(\by_i^j,(-1)^\sigma \bn_i^j)}(\by, \bn),
\end{equation}
with index $j$ running over points within a fiber, $i$ running over all fibers and $\sigma$ taking care of including forward and backward orientations. We use the same evolution process as in Eq. (\ref{eq:contenh}) in which $F = F_{\Gamma}$ now serves as initial condition, to create a diffused density $(\by,\bn) \mapsto W_{F}(\by,\bn,t)$:

\begin{equation}\label{eq:contenhF}
\left\{
\begin{aligned}
    &\partial_t W_{F}(\by,\bn,t) = \left( D_{33}(\bn \cdot \nabla_{\by})^2 + D_{44} \Delta_{S^2} \right) W_{F}(\by,\bn,t), \\
    & W_F(\by,\bn,0) = F(\by,\bn).
  \end{aligned} \right.
\end{equation}
We solve the system in (\ref{eq:contenhF}) by convolution with the corresponding kernel, recall Fig. \ref{fig:kernelvis}, and call this the \emph{local} FBC (LFBC):

\begin{equation}\label{eq:ptconvF}
\LFBC(\cdot,\Gamma) = W_F(\cdot,t)=(p_t \ast_{\mathbb{R}^3 \rtimes S^2} F)(\cdot),
\end{equation}
with the shift-twist convolution as given in Eq. (\ref{eq:shifttwistconv}). This is illustrated in Fig. \ref{fig:spurfibschematic} in the 2D case. We can now define the FBC for fiber $\gamma_i$ with respect to the bundle $\Gamma$ as the integral of this density along the fiber:

\begin{equation}
\FBC(\gamma_i,\Gamma) = \frac{1}{l_i} \int_0^{l_i} \LFBC(\gamma_i(s),\Gamma) \; \rmd s.
\end{equation}
This results in a global property of the fiber, but spurious fibers often only locally deviate from the bundle as in Fig. \ref{fig:spurfibschematic}. To this end, we compute for each fiber the minimum of such integrals along the fiber over intervals of length $\alpha$:

\begin{equation}
\FBC^{\alpha}(\gamma_i,\Gamma) = \min_{a \in [0,l_i-\alpha]} \frac{1}{\alpha} \int_a^{a+\alpha} \LFBC(\gamma_i(s),\Gamma) \; \rmd s.
\end{equation}
The parameter $\alpha$ defines the scale over which spuriousness of fibers can be detected and is much smaller than the average fiber length. Our primary interest is not the $\FBC^{\alpha}$ value itself, but rather how it compares to the average coherence of fibers in the bundle, so finally we define the \emph{relative} fiber to bundle coherence (RFBC) as:

\begin{equation}
\RFBC(\gamma_i,\Gamma) = \frac{\FBC^{\alpha}(\gamma_i,\Gamma)}{\AFBC(\Gamma)}.
\end{equation}
Here $\AFBC(\Gamma)$ is the \emph{average} fiber to bundle coherence indicating the overall coherence of the $N$ fibers in the bundle $\Gamma$, defined as

\begin{equation}
\AFBC(\Gamma) = \frac{1}{N} \sum_{i=1}^N \FBC(\gamma_i,\Gamma).
\end{equation}
To summarize, the $\RFBC(\gamma_i,\Gamma)$ of a fiber $\gamma_i$ in a bundle $\Gamma$ is a measure for how well aligned the least aligned part of $\gamma_i$ is, compared to the average coherence of the total bundle.

In practice, we evaluate the convolution in Eq. (\ref{eq:ptconvF}) only in the fiber points. We compute the $\LFBC(\gamma_i^k,\Gamma)$, the diffused density in the oriented point $\gamma_i^k = (\by_i^k, \bn_i^k)$, recall the notation in (\ref{eq:notationgamma}), as follows:

\begin{equation}
\LFBC(\gamma_i^k,\Gamma) = \frac{1}{N_{tot}}\sum_{\sigma=1}^2 \sum_{j=1}^N \sum_{q=1}^{N_j} p_t \left(\bR^T_{(-1)^{\sigma} \bn_j^q}(\by_i^k - \by_j^q), \bR^T_{(-1)^{\sigma} \bn_j^q} \bn_i^k\right),
\end{equation}
where $\bR_{\bn_j^l}$ is any rotation matrix such that $\bR_{\bn_j^l} \be_z = \bn_j^l$, index $q$ sums the contributions along a fiber, index $j$ runs over all the fibers and $\sigma$ as before. The $\FBC^{\alpha}$ can then be computed as follows:

\begin{equation}
\FBC^{\alpha}(\gamma_i,\Gamma) = \min_{a \in [0,N_i-\alpha]} \frac{1}{\alpha} \sum_{k=a+1}^{a + \alpha} \LFBC(\gamma_i^k,\Gamma),
\end{equation}
where $a, \alpha \in \mathbb{N}$ in this discrete case, so the LFBC is summed along short intervals of the fiber. Likewise, the AFBC can be computed as

\begin{equation}
\AFBC(\Gamma) = \frac{1}{N} \sum_{i=1}^N \frac{1}{N_i} \sum_{k=1}^{N_i} \LFBC(\gamma_i^k,\Gamma).
\end{equation}
We apply this method in Section \ref{se:ORexperiment} for quantifying the coherence of tractography results of the optic radiation and classifying the spurious fibers.

\section{Experiments and Results}\label{se:results}
In this section we extensively test the performance of our CSD enhancement method (A) and the FBC method (B), recall Fig. \ref{fig:csdenhpipeline} and Sections \ref{se:enhancement} and \ref{se:coherence}, in three different experiments:

\begin{itemize}
\item We use the HARDI Reconstruction Challenge dataset \cite{Daducci2013}, which is artificial data with known ground truth, to quantitatively evaluate the CSD enhancement method (A) on deterministic tractography in Section \ref{se:HARDIrecon}.
\item In Section \ref{se:evaluationDWMRI} we show on DW-MRI human brain data that the enhancement (A) have a positive effect on deterministic tractography, for different acquisition protocols of the data. Furthermore, on this DW-MRI dataset and on the phantom dataset we compare our method to previous work \cite{Duits2013}, where a DTI-based FOD is used in combination with nonlinear PDE flow.
\item In the third and last experiment, we reconstruct the optic radiation in human clinical data, see Section \ref{se:ORexperiment}. We include an extensive evaluation of our methods, the enhancement of the FOD (A) and the use of the FBC to classify and remove spurious fibers (B), and the combination of both methods. We show that the reproducibility of the probabilistic tractography has increased, resulting in a more stable localization of the tip of the Meyer's loop.
\end{itemize}    

For all datasets Mathematica \cite{Mathematica10} was used to perform the contour enhancement algorithm and the CSD, which in practice produces the same results as the MRtrix CSD implementation when the same deconvolution kernel is used. MRtrix software  \cite{Tournier2012} was used to perform fiber tractography. The coherence quantification was implemented in C++. In Section \ref{se:HARDIrecon} we make use of the Tractometer \cite{cote2012tractometer,Cote2013} (\url{http://www.tractometer.org/}) to evaluate tractography results. Visualization was done in either the FiberNavigator (\url{https://github.com/scilus/fibernavigator},\cite{chamberland2014real}), Mathematica, or the open source vIST/e tool (Eindhoven
University of Technology, Imaging Science \& Technology Group, \url{http://bmia.bmt.tue.
nl/software/viste/}).\\

\subsection{HARDI Reconstruction Challenge}\label{se:HARDIrecon}
\begin{figure}[t!]
        \centering        
                \includegraphics[width=\textwidth]{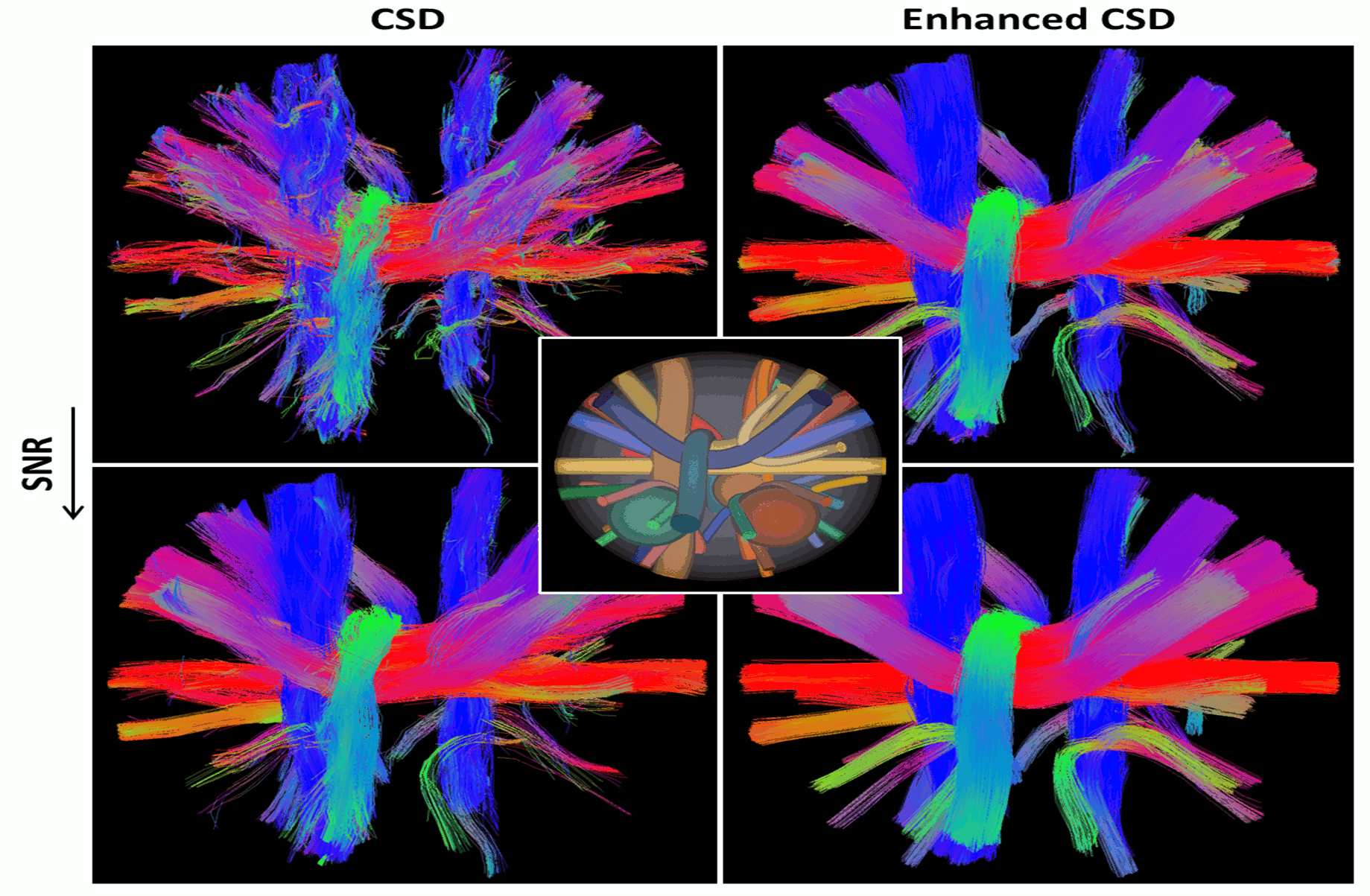}
        \caption{{\bf Tractrography results on the ISBI Challenge dataset.}
        Deterministic tractography results of CSD (left) and enhanced CSD (right) with $SNR=2$ (top) and $SNR=4$. The colors correspond to the direction of the fibers. The dataset consists of crossing, branching and kissing fiber bundles. The tractography on enhanced CSD results in better aligned fibers and a fuller reconstruction of the bundles. The ground truth configuration of the bundles is depicted in the center.}\label{fig:isbifull}
\end{figure}

The following experiment is performed on a digital phantom dataset that was designed for the ISBI 2013 Reconstruction Challenge \cite{Daducci2013,Caruyer2014}. It is used in combination with the Tractometer \cite{cote2012tractometer,Cote2013}, as a benchmark to compare different reconstruction  and tracking methods. The phantom is inspired by the Numerical Fiber Generator \cite{Close2009} and the code to reproduce it is freely available as part of the Python package Phantomas (\url{http://www.emmanuelcaruyer.com/phantomas.php}). This synthetic dataset is of size $50\times 50 \times 50$ voxels with a resolution of $1\times 1 \times 1$ mm$^3$. It consists of $27$ simulated white matter bundles, designed to resemble challenging branching, kissing and crossing structures at angles between $30$ and $90$ degrees, with various curvature and bundle diameters ranging from $2\,\textrm{mm}$ to $6\,\textrm{mm}$.  An image indicating the ground truth fiber configuration is shown in the centre of Fig. \ref{fig:isbifull}. 
 
The idea behind the signal simulation is that every voxel is subdivided into multiple sub-voxels, each one with its own attenuation profile. The final signal arrives from integrating the contribution of all the sub-voxels. Then, it is possible to combine multiple compartment types in every voxel with added Rician noise. This allows for modelling complex configurations as well as taking into account partial volume effects. While the Numerical Fiber Generator uses a tensor-like model to simulate the signal in the sub-voxels, Phantomas uses a CHARMED-based model \cite{Assaf2005}. The CHARMED model based on the Söderman-Jönsson cylinder model \cite{SODERMAN199594} captures well the non-Gaussian behaviour of the diffusion signal for large b-values. The main reason why we selected the ISBI phantom is that it is linked with the Tractometer that allows for performing quantitative evaluations of the tractography results, using global metrics as demonstrated in the subsequent experiments.

For the experiments presented in this section we used 64 uniformly distributed gradient directions using a $b$-value of $3000\,\textrm{s/mm}^2$ with different signal to noise ratios (SNRs). We use spherical harmonics in CSD with maximal order 8, resulting in $45$ estimated coefficients on each position. We then enhance the resulting FOD functions using our contour enhancement algorithm with varying parameters. From the evolutions described in Eq. (\ref{eq:contenh}) we see by a basic rescaling argument that it is sufficient to vary $t$ and the ratio $D_{33} / D_{44}$. The larger this ratio, the more preference the spatial diffusion gets over the angular diffusion, resulting in elongated kernels (visualized by thin glyphs). A smaller ratio $D_{33}/D_{44}$ is better suited in regions where the curvature of bundles is higher (visualized by thicker glyphs). The higher the diffusion time $t$, the more context is taken into account. When $t$ is too large, fiber bundles with high curvature can be damaged or false positives could be created. Taking this into consideration, we choose our parameters as follows: we fix spatial diffusivity parameter $D_{33}= 1.0$, we take the angular diffusivity parameter $D_{44} \in \{0.005,0.01,0.02,0.04\}$ and diffusion times $t \in [0,5]$.

Tractography results for the entire dataset are shown in Fig. \ref{fig:isbifull}. We can recognize the positive effect of the enhancements on deterministic tractography: we see less dropouts, better aligned fibers and better continuation of fibers at crossings. An extensive quantification of the performance of our method is done at the voxel level using the FODs and at the macroscopic level using tractography in Sections \ref{se:localmetrics} and \ref{se:globalmetrics}, respectively. Both sections support the results summarized in Fig. \ref{fig:isbiresults}.

\subsubsection{Local Metrics}\label{se:localmetrics}
 We compare reconstructed FODs locally with the ground truth using only the orientation of the peaks. Let $M$ be the set of voxels $v$ in the white matter mask, then we denote the ground truth number of peaks in a voxel $v$ by $N^{v}$ and the orientations corresponding to the peaks by $\bn^v_{i,\text{true}}$, $i = 1, \dots,N^{v}$. 

 Maxima of the constructed FOD are found by evaluating the FODs on a $60^{\text{th}}$ order icosahedron tessellation with $18606$ antipodally symmetric points, giving an angular resolution of less than $1$ degree. Maxima are taken into account only if it exceeds a threshold of $0.1$, the same value we use as threshold in the tractography. Let $O^v_{\text{est}}$ be the set of peak orientations in voxel $v$ estimated from the FOD. The average angular error in degrees can then be computed by:

\begin{equation}
 \vartheta = \dfrac{ \underset{v \in M} \sum \: \underset{i=1}{\overset{N^{v}} \sum} \: \underset {\bn \in O^v_{\text{est}}} \min \dfrac{180}{\pi} \arccos(|\bn^v_{i,\textrm{true}} \cdot \bn|)} { \underset{v \in M}\sum N^{v}}.
\end{equation}

\begin{figure}[t!]
        \centering
                \includegraphics[width=\textwidth]{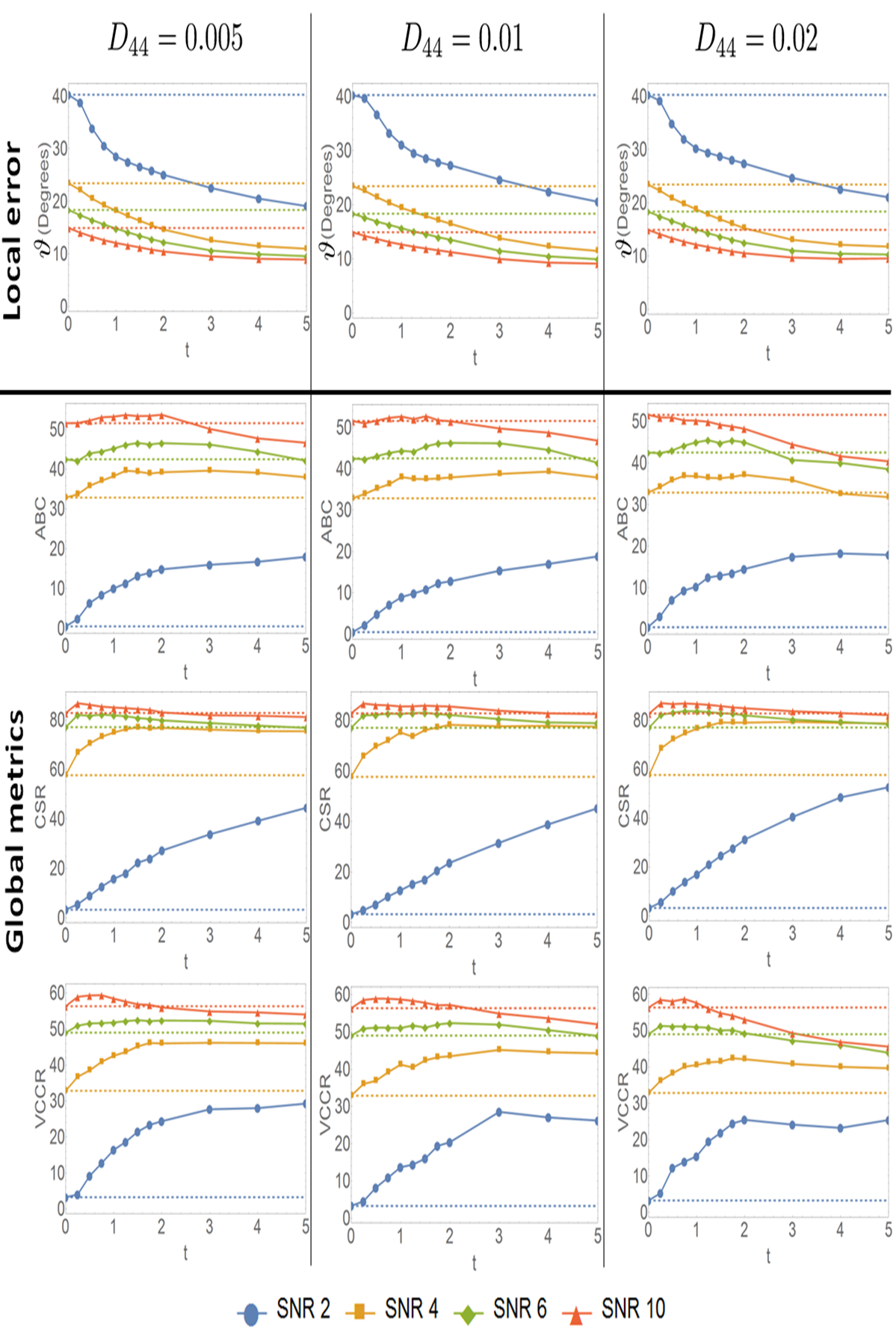}
        \caption{{\bf Quantitative evaluation of the effect of enhancements.}
        Evolution of the local error and global metrics for three different choices of $D_{44}$ and four different SNRs as we increase the diffusion time $t$. The top row shows the average angular error of the FOD peaks, the rows below show the average bundle coverage (ABC), connection to seed ratio (CSR) and the valid connection to connection ratio (VCCR), computed from the tractography results.}\label{fig:isbiresults}
\end{figure}

In the top row of Fig. \ref{fig:isbiresults} we show the effects contour enhancement for different ratios of $D_{33}$ and $D_{44}$ upon variation of the diffusion time. The results are given for substantially low SNR levels $10$, $6$ and $4$ and $2$. These SNRs are computed w.r.t. the non-DW image. Specifically, if the b=0 intensity is 1 then the standard deviation of the Rician noise distribution is 1/SNR. In all cases a clear improvement is found compared to CSD without enhancements and the more noise, the more the angular error is decreased. Higher diffusion times give better results and around $t=5$ the angular error is almost stable. It can also be seen that the combination of CSD with enhancements at lower SNRs gives lower angular errors than just CSD for the higher SNRs.

 There is no significant difference in the FODs between the different $D_{44}$ values. Even though it is visible that more angular diffusion leads to fatter glyphs, for the orientation of the peaks the precise value of $D_{44}$ is not of great importance: the angular errors for $D_{44} = 0.005$ are slightly smaller, but there is not much difference with the higher values of $D_{44}$. \\

\subsubsection{Global Metrics}\label{se:globalmetrics}

At the macroscopic level we are interested in the impact of the enhanced local reconstruction on the quality of the global connectivity. The deterministic MRtrix tractography is used as described in Section \ref{se:tractography}, with seeds randomly selected in the white matter mask. The tracks have a minimum length of $10$ mm and new seed points are chosen until 10000 streamlines are selected. For every FOD, the tractography is repeated five times with the same settings, to average out the variability in the tracking algorithm output. We then use the Tractometer \cite{Cote2013} to perform a fiber tracking analysis based on the ground truth and the five results are averaged. The Tractometer outputs values for various metrics, from which we use the Valid Connections (VC), Invalid Connections (IC) and No Connections (NC). They indicate the percentage of tracks that correctly connect, incorrectly connect or do not connect gray matter areas in the dataset, respectively. We also use the Average Bundle Coverage (ABC), the percentage of voxels in a bundle that is crossed by a valid streamline, averaged over all bundles. We combine the (VC), (IC) and (NC) in two metrics introduced in \cite{Girard2014}:
\begin{itemize}
\item Connection to Seed Ratio (CSR), which represents the probability that a generated fiber actually connects two gray matter areas, computed as $100 \% -$NC.
\item Valid Connection to Connection Ratio (VCCR), the probability that a connecting fiber is correct, computed as VC/(VC+IC).
\end{itemize}

The results for the ABC, CSR and VCCR with the same enhancement parameters and SNRs as for the local metric are given in Fig. \ref{fig:isbiresults}. Similar remarks hold for the global metrics as for the angular error. For all three metrics and all SNRs the enhancements lead to an improvement compared to CSD, the only exception being the ABC for $SNR = 10$ and $D_{44}=0.02$. Furthermore, as the SNR decreases, the larger diffusion times are beneficial and the more significant the improvement is. The best results are obtained for $D_{44} = 0.005$. We expect that truncation of the spherical harmonics already introduces some angular smoothing of the FODs on this artificial dataset, explaining the small effect of $D_{44}$ in the experiments. Furthermore, we see that the diffusion time $t$ truly acts as a regularization parameter, resulting in a robustness for the metrics with respect to the SNRs: the higher the diffusion time, the smaller the differences in the metrics between the different SNRs.

Seeding from the white matter voxels can lead to an over-representation of the number of fibers in longer fiber bundles with respect to the shorter bundles \cite{smith2013sift}. The longer bundles thereby have a larger contribution to the global metrics than the shorter bundles, which could lead to an overestimation of the fiber bundles. As proposed in \cite{smith2012anatomically}, we compared the global metrics when seeding from the gray/white matter interface for CSD and one specific set of enhancement parameters. The global metrics for that seeding strategy were slightly lower for CSD and comparable when including enhancements. For the sake of comparing our enhancement method with CSD, we therefore believe it is fair to use seeding from the white matter mask.

\begin{figure}[t!]
        \centering
        \includegraphics[width=\textwidth]{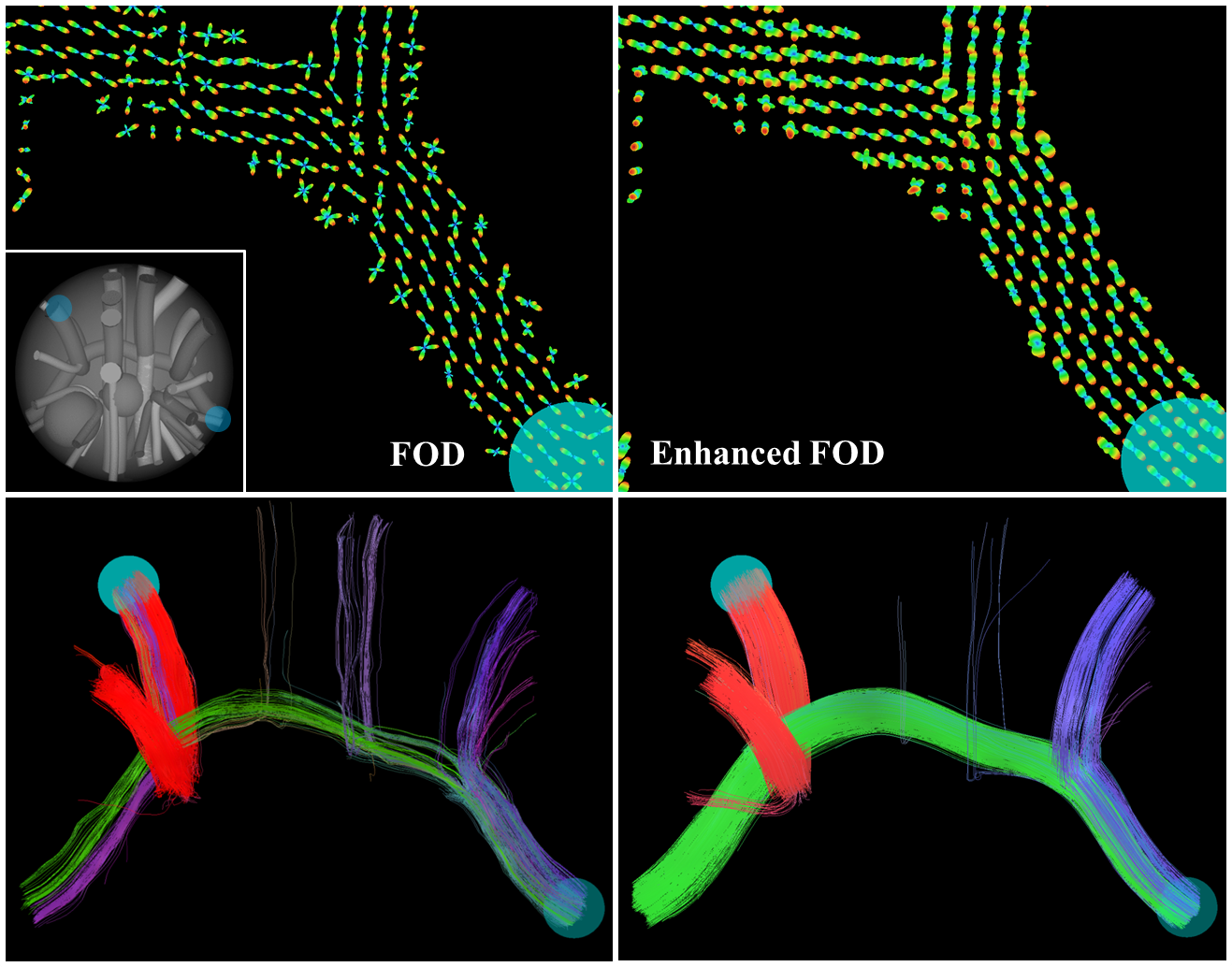}       
        \caption{{\bf Comparison of CSD and enhanced CSD on the ISBI dataset.}
        The top row shows that glyphs visualizing the FOD are better aligned, especially at the boundary of the bundle. In this case, color corresponds to the radius of the glyph. The bottom row depicts the tractography results, showing only the streamlines that pass through the indicated spheres. Here the $SNR = 4$ and the parameters used for the enhancements are $D_{33} = 1.0$, $D_{44} = 0.02$, $t = 4.0$. The ground truth image with the same viewpoint as the bottom figures is depicted on the left.}\label{fig:prtscrns}
\end{figure}

The convincing improvement in the global metrics is supported by Fig. \ref{fig:prtscrns}, that shows a selection of the fiber bundles in the dataset. It can be seen that after enhancements, there are more valid connections in the green bundle and less wrong exits in the red bundle, leading to a higher (VCCR) and a better bundle coverage. The glyphs in the top row show that the enhancements improve alignment of glyphs, especially at the boundary of the fiber bundles, where the original CSD result tends to suffer from partial volume effects.

\subsection{Evaluation and Comparison on DW-MRI Data}\label{se:evaluationDWMRI}
In this experiment we consider a DW-MRI dataset of a part of a human brain, previously used in \cite{Duits2013}. The study was approved by the local ethical commitee of Maastricht University, and informed written consent was obtained from the subject. Although the dataset consists of only 10 axial slices, the corpus callosum, corona radiata and superior longitudinal fasciculus are (partly) present in the data. We show that the combination of CSD and enhancement is well-suited for different combinations of the $b$-value and the number of gradient directions used in the acquisition. Furthermore, we make a qualitative comparison with the DTI-based method of \cite{Duits2013} on this dataset and conclude with a brief quantitative comparison with this method on the dataset of \ref{se:HARDIrecon}.

\subsubsection{Robustness with Respect to the Acquisition Parameters}
The acquisition was performed on a 3T Siemens Allegra scanner, with FOV 208x208mm and voxel size 2x2x2mm. During the data acquisition, a brain region consisting of 10 axial slices was scanned with the following combinations of $b$-values and $N_o$, the number of orientations: $b = 1000$ s/mm$^2$ with $N_o = 49$, $b = 1000$ s/mm$^2$ with $N_o = 121$ and $b = 4000$ s/mm$^2$ with $N_o = 49$. We use again CSD with spherical harmonics up to order 8. The higher $b$-value is obtained by using a stronger gradient pulse, making the acquisition more sensitive to detail in the tissue structure, but also inducing a lower SNR. Increasing the number of gradient directions gives a better angular resolution. We use deterministic tractography, with three seed regions manually selected in the middle of the corpus callosum, corona radiata and superior longitudinal fasciculus.

\begin{figure}[t!]
        \centering        
        \includegraphics[width=\textwidth]{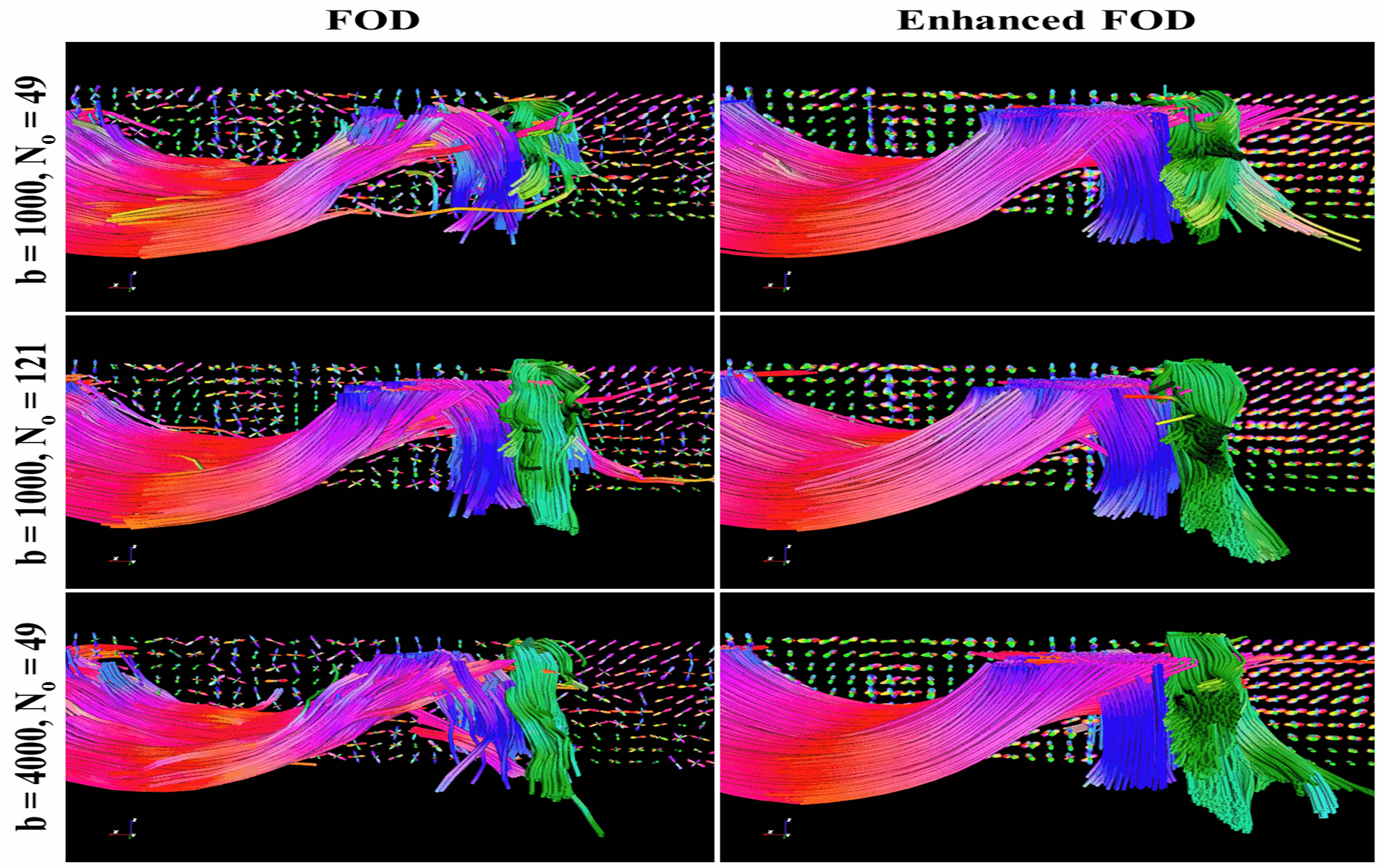}       
        \caption{{\bf Comparison between CSD and enhanced CSD of tractography on human data.}
        The tractography results on CSD and enhanced CSD data of the corpus callosum (mostly red), coronia radiata (mostly blue) and superior longitudinal fasciculus (mostly green), with the color related to the fiber direction. Enhancements are performed with $D_{33}=1.0$, $D_{44}=0.02$, $t = 4.0$. All three bundles are more apparent after enhancements and more fibers pass the crossings.}\label{fig:trackingcomparison}
\end{figure}

In the right column of Fig. \ref{fig:trackingcomparison} we show that after enhancements, the FOD allows for a more coherent reconstruction of the three bundles. Especially in the region where the three bundles come together, it can be seen that the fibers have a better propagation through the crossings. Moreover, the FODs after enhancements are very similar to each other, visible in the glyph visualization, leading to three tractography results supporting similar fiber bundles. This is an improvement with respect to CSD without enhancement, shown in the left column of Fig. \ref{fig:trackingcomparison}. There we find more noisy FODs with more variation between the different protocols. This is also reflected in the tractography results, that contain more spurious fibers than after the enhancements. 

We conclude, just like in the first experiment on the phantom data, that applying enhancements induces more robust tractography also on real DW-MRI data, in this case in the sense that it is less sensitive to the acquisition parameters $b$ and $N_o$.

\subsubsection{Comparison with a DTI-based FOD}
In the next experiment we compare the performance of our combination of CSD with enhancements with the method in \cite{Duits2013} which proposed to combine DTI with \emph{non-linear} PDE-based enhancement obtained from successively applying erosions and diffusions. Let us briefly describe this method, for details we refer to \cite{Duits2013}, and an implementation of the PDE enhancements can be found in the HARDI package for Mathematica available at (\url{http://bmia.bmt.tue.nl/people/RDuits/HARDIAlgorithms.zip}). First an FOD on positions and orientations that we call $U_{DTI}$ was constructed via a transformation of the tensor field $D$ fitted to the data \cite{Basser1994}, according to the following definition \cite{aganj2011hough,Duits2013}:

\begin{equation}
        U_{DTI}(\by,\bn) = \frac{1}{4 \pi \int_{\Omega} \sqrt{\det(D(\bx))}d \bx} . (\bn^T D^{-1}(\by) \bn)^{-\frac{3}{2}}.
\end{equation}
This FOD is then sharpened with PDE erosions, a type of morphological enhancement adapted from \cite{Burgeth2009b}, on $\mathbb{R}^3 \rtimes S^2$ and regularized with nonlinear diffusions to find crossing structures from DTI. 

\begin{figure}[ht!]
        \centering
                \includegraphics[width=\textwidth]{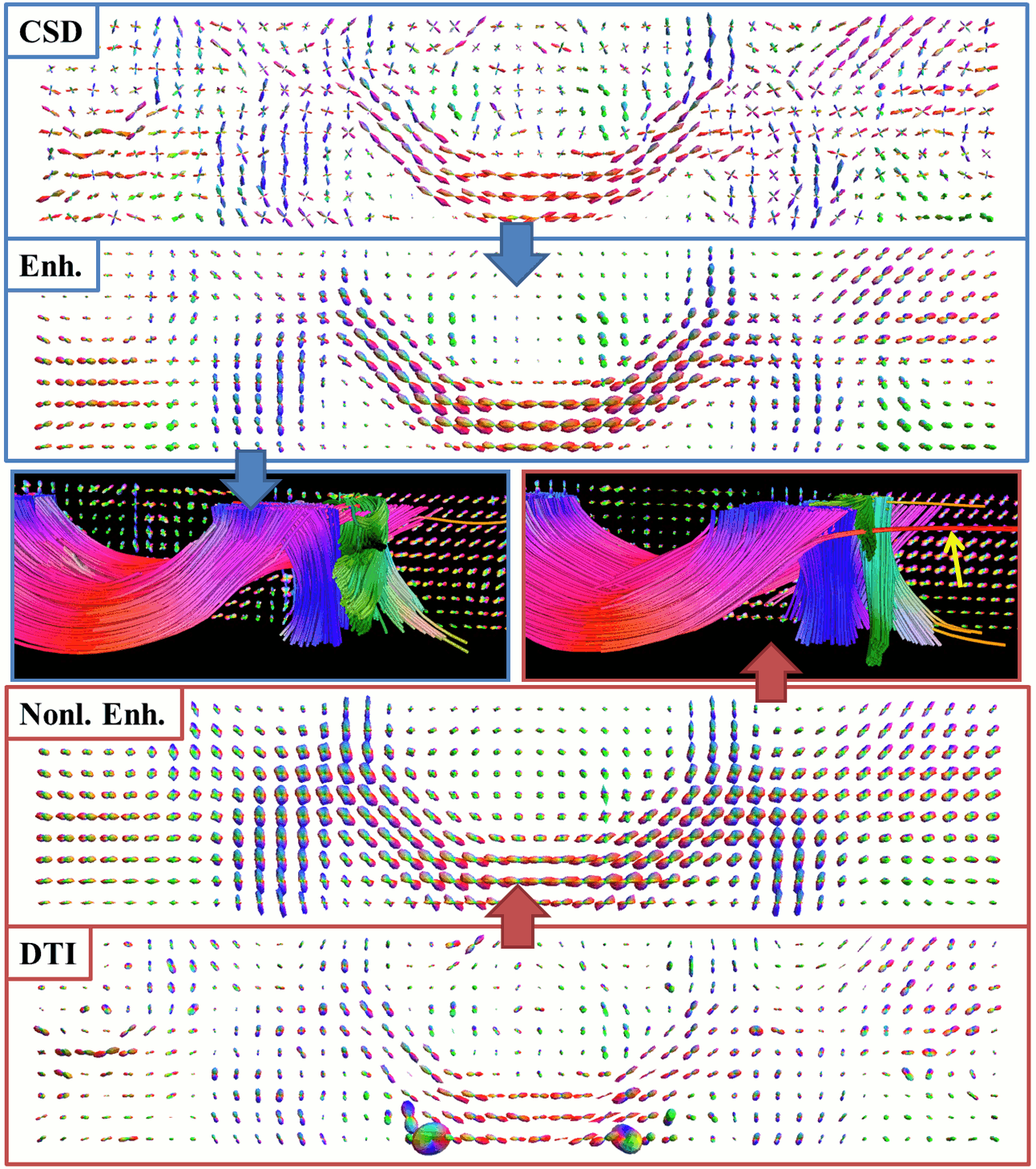}        
        \caption{{\bf Comparison of glyph fields and tractography results between enhanced CSD and a DTI-based FOD.}
        Glyph visualization of an axial slice of a dataset supporting the presence of the corpus callosum (mostly red), corona radiata (mostly blue) and superior longitudinal fasciculus (mostly green). Contour enhancement for CSD is performed with $D_{33} = 1.0, D_{44} = 0.02, t =4$. Erosions and nonlinear diffusions for the DTI-based method are done with parameters as in \cite{Duits2013}. The tractographies corresponding to the two methods are shown in the middle. Outliers such as the red fiber, indicated by the arrow, occur due to the use of high regularization coefficients.}\label{fig:glyphcomparison}
\end{figure}

Previously in \cite{Duits2013}, the same dataset as in Fig. \ref{fig:trackingcomparison} for acquisition parameters $b=1000$ s/mm$^2$ and $N_o = 49$ was processed. Here we compare the FOD obtained with CSD, that we call $U_{CSD}$ here, with $U_{DTI}$ in the top and bottom figures, respectively, of Fig. \ref{fig:glyphcomparison}. Unlike DTI, which is limited by the Gaussian assumption of the diffusion profile, CSD can estimate multiple fiber orientations within a voxel. Furthermore, we see that the large glyphs in the Centrum Semiovale in the bottom figure are not apparent in $U_{CSD}$. Applying (linear) enhancements, as explained in Section \ref{se:enhancement}, to $U_{CSD}$ gives the second figure, and the approach in \cite{Duits2013} using erosions/(nonlinear) enhancements applied to $U_{DTI}$ gives the second figure from below. It can be seen that also the enhanced DTI glyphs supports multiple fiber directions within voxels via extrapolation \cite{Vesna2010,Duits2013}, but at the cost of high regularization. Another noticeable difference is the fact that the glyphs in the CSD case are slimmer and crossings are more clearly defined. Whether two separate maxima are visible at a crossing is less dependent on the diffusion parameters in the PDE diffusion.

Besides the visual comparison of the FOD glyphs, we provide deterministic tractography results for both procedures in the middle of Fig. \ref{fig:glyphcomparison}. It can be observed that both methods produce reasonable results, although the one obtained from the enhanced DTI dataset seems oversmoothed and outliers (indicated with the yellow arrow) can occur. This is due to the extreme diffusion parameters needed to perform the FOD extrapolation. We find that visually the combination of CSD and linear enhancements yields better tractography than DTI combined with erosions and nonlinear enhancements.

To provide a more quantitative and complete comparison of DTI, DTI and nonlinear enhancements, CSD and CSD with linear enhancements, we also include results of the experiment in Section \ref{se:HARDIrecon} for the DTI methods, see Table \ref{tab:isbiresults}. We heuristically determined good parameter settings for the nonlinear enhancement of DTI: erosions \cite[Eq. (59)]{Duits2013} with $D_{11} = 0.5$, $D_{44} = 0.2$, $t = 2.0$ and diffusion \cite[Eq. (55)]{Duits2013} with $D_{11} = 0.2$, $D_{33} = 1.0$, $D_{44} = 0.02$ and $t = 3$. In Table \ref{tab:isbiresults} is shown that applying enhancements for contextual regularization of the FOD is beneficial for both DTI and CSD. The lower the SNR, the more evident the improvements become. Furthermore, we see that in terms of the local metric, the angular error $\theta$ of the peak orientations, the DTI methods can compete with the CSD based methods. However, the global metrics are significantly higher for CSD based methods. The quantitative results on the phantom data in Table \ref{tab:isbiresults} are in line with the qualitative comparison on real data in Fig. \ref{fig:glyphcomparison}.

\begin{table}[htbp]
  \centering
  \caption{For two SNR values, the results are shown for the DTI method described in Section \ref{se:evaluationDWMRI}, with or without nonlinear enhancements. We compare with CSD and a specific instance of enhanced CSD with parameters $D_{33} = 1$, $D_{44} = 0.01, t = 2$. For local metric $\theta$ lower is better, for the other metrics higher is better. In boldface are the best results for the DTI and CSD methods.\\}

    \begin{tabular}{|l|ll|ll|}

    \hline
     \textbf{SNR 4}  & DTI   & DTI enh & CSD   & CSD enh \\
     \hline
    \textbf{$\theta$} (deg.) & 33.9  & $\mathbf{15.2}$  & 23.4  & $\mathbf{16.3}$ \\
    \textbf{ABC} (\%) & 14.3  & $\mathbf{18.1}$  & 32.9  & $\mathbf{37.9}$ \\
    \textbf{CSR} (\%) & 50.2  & $\mathbf{54.1}$  & 57.6  & $\mathbf{78.2}$ \\
    \textbf{VCCR} (\%) & 17.5  & $\mathbf{20.0}$  & 32.9  & $\mathbf{43.5}$ \\
    \hline
         
    \textbf{SNR 10}      & DTI   & DTI enh & CSD   & CSD enh \\
    \hline
    \textbf{$\theta$} (deg.) & 23.8  & $\mathbf{13.0}$  & 14.9  & $\mathbf{11.1}$ \\
    \textbf{ABC} (\%) & 15.5 & $\mathbf{19.9}$  & $\mathbf{51.6}$  & 51.5 \\
    \textbf{CSR} (\%) & $\mathbf{69.1}$  & 64.6  & 82.8  & $\mathbf{85.5}$ \\
    \textbf{VCCR} (\%) & 17.1  & $\mathbf{24.3}$ & 56.4  & $\mathbf{57.2}$ \\
    \hline
    \end{tabular}%
  \label{tab:isbiresults}%
\end{table}%

\subsection{Improved Reconstruction of the Optic Radiation}\label{se:ORexperiment}
The optic radiation (OR) is a white matter fiber bundle connecting the primary visual cortex and the lateral geniculate nucleus (LGN), see Fig. \ref{fig:prtscrn_OR}. The most anterior part of the OR is called the Meyer's loop (ML), of which the exact location is of interest for treatment of temporal lobe epilepsy \cite{falconer1963follow,Sherbondy2008Contrack,Tax2014,Meesters2013}. During neurosurgery, a part of the temporal lobe is resected. To ensure that the OR remains intact to prevent visual field defect, it is crucial to know the distance from the tip of the Meyer's loop to the Temporal Pole (ML-TP) \cite{powell2005mr}, which shows large interpatient variability \cite{nilsson2007intersubject}. 

 We use DW-MRI scans of four subjects, performed on a 3.0T Philips Achieva MR scanner, with $b=1000$ s/mm$^2$, $N_o=32$ and a spatial resolution of 2x2x2 mm. All subjects gave written informed consent; the study was approved by the Medical Ethics Committee of Maastricht University Medical Center (N 43386.068). The data is acquired from healthy volunteers, and ground-truth ML-TP distance is not known. Therefore accuracy of this measure of our methods cannot be checked, instead we focus on consistency and reproducibility. We apply CSD to the data to construct the FOD, with spherical harmonics up to order 6 requiring the estimation of 28 coefficients (as 32 directions are insufficient to estimate the 45 coefficients when a spherical harmonic order 8 is used, when not using super-resolution as in \cite{Tournier2007}). We seed from the LGN and include all fibers that reach the primary visual cortex. Both regions of interest are selected manually on a T1-weighted image. We use probabilistic fiber tracking as described in Section \ref{se:tractography}. 

 We demonstrate the effect of the enhancement of CSD and the use of the FBC measure in Sections \ref{se:methodA} and \ref{se:methodB}, respectively, in this relevant clinical setting. A quantitative comparison of the four methods CSD (O), CSD + enhancement (A), CSD + FBC (B) and CSD + enhancement + FBC (A+B) is provided in Section \ref{se:quantitativeOR}.  We show that the enhancement and/or the removal of spurious fibers, but in particular the combination of both methods, allows for a more stable computation of the ML-TP distance than the original tractography result.

\subsubsection{Effect of the Enhancement of CSD on Tractography of the OR}\label{se:methodA}
In this section, we apply the PDE enhancement (step A) to the CSD FOD as before, with parameter settings $D_{33} = 1$, $D_{44} = 0.01$ and $t=2$. After the enhancement we apply the sharpening deconvolution transform \cite{descoteaux2009deterministic} and probabilistic tractography with 10000 streamlines. We compare the results of the tractography on the subjects both before and after the enhancement in Fig. \ref{fig:OR4subjects}. We see that the tracking on enhanced data generally shows less spurious fibers, and has a better pronounced tip of the Meyer's loop. However, the optic radiation is a highly curved structure, where the advantage of the enhancement of elongated structures cannot be fully exploited. To further reduce the spurious fibers, we explore our other approach in the next section.

\subsubsection{Effect of the FBC measure on Tractography of the OR}\label{se:methodB}
In this section, we apply probabilistic tractography on subject 1, with 20000 streamlines and including state of the art data scoring as in \cite{Tax2014} (only relying on the data term, i.e. $\lambda = 0$ in \cite[Eq.(11)]{Tax2014}), see Fig. \ref{fig:ORcomparison}.

The kernel parameters for the coherence quantification (step B) are set to $D_{33} = 1$, $D_{44} = 0.04$ and $t = 1.4$ for the convolution \cite{Duits2013}. Let $\Gamma$ be the set of the 1000 most anterior fibers in a tractography of the OR, that roughly form the Meyer's loop. We compute the LFBC and subsequently the RFBC for all the fibers in $\Gamma$. 

Then we take $\epsilon_{max}(\Gamma) := \underset{\gamma \in \Gamma} \max \: \RFBC(\gamma,\Gamma)$, the RFBC corresponding to the ``central'' fiber, in the sense that it is most coherent with the fiber bundle. We define the filtered set $\Gamma_{\epsilon}$ as

\begin{equation}
\Gamma_{\epsilon} := \left\{\gamma \in \Gamma \: \vline \: \RFBC(\gamma,\Gamma) \geq \epsilon \right\}, \qquad 0 \leq \epsilon \leq \epsilon_{max}.
\end{equation}
This means the parameter $\epsilon$ acts as a threshold parameter and can be set such that fibers with a high spuriousness are removed. The fiber point in $\Gamma_{\epsilon}$ that is closest to the temporal pole defines the ML-TP distance. We repeat the probabilistic tractography five times with the same settings on the same data, to qualitatively compare different stochastic realizations of the tractography method. The original OR reconstructions are shown in the top row of Fig. \ref{fig:ORcomparison}. We observe that due to the presence of spurious fibers, the tip of the Meyer's loop (indicated by the orange spheres) is estimated at different locations. When we set the threshold $\epsilon = 0.1 \epsilon_{max}$, removing in these cases between $6$\% and $8$\% of the most spurious fibers, we obtain the results as shown in the bottom row of Fig. \ref{fig:ORcomparison}. It can be seen that the resulting fiber bundles are very similar to each other, demonstrating less variation in the localization of the tip. 

\subsubsection{Quantitative Comparisons on Four Subjects}\label{se:quantitativeOR}
\begin{figure}[t!]
        \centering
                \includegraphics[width= \textwidth]{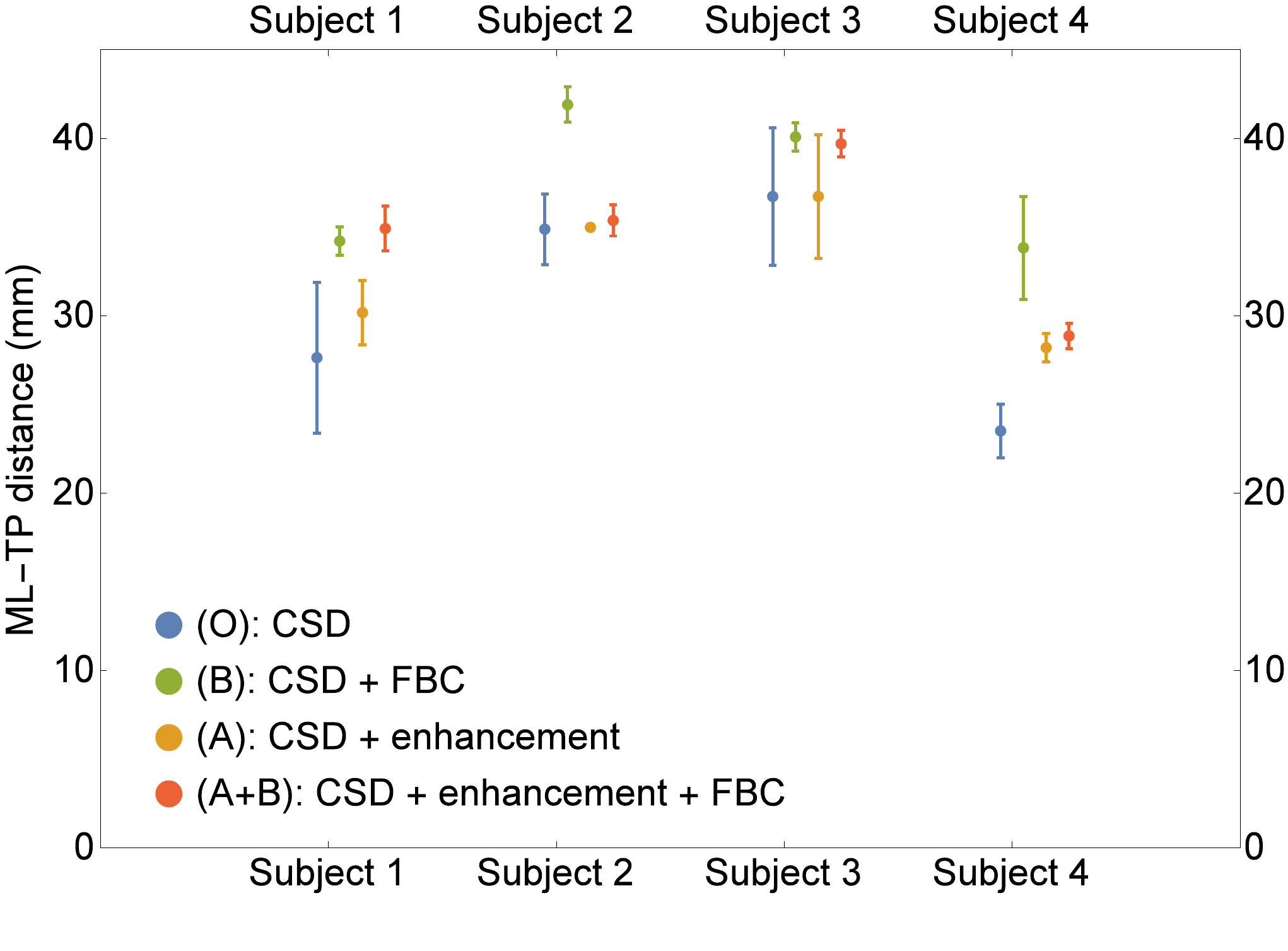}
        \caption{{\bf Boxplots of the ML-TP distances.} For the four subjects, we show the mean of the ML-TP distance over ten tractography results, plus two standard deviations. The four different methods are indicated with different colors. The combination CSD + enh + FBC is the most robust in producing stable results.}\label{fig:MLTP_distances}
\end{figure}

To support our claims of the two previous sections, we test the effect of our methods on the stability of the ML-TP distance under different stochastic realizations. Here we perform probabilistic tractography with 10000 fibers ten times with the same settings, for each of the four subjects and each of the four methods (CSD, CSD + enh, CSD + FBC and CSD + enh + FBC). The FBC measure is computed from the 1000 most anterior fibers as in the previous experiment and the threshold is set to $\epsilon = 0.05 \epsilon_{max}$. We compare the mean ML-TP distance and sample standard deviation determined from the tracking results of each of the methods. The results are summarized in the boxplots in Fig. \ref{fig:MLTP_distances}. The figure strongly supports the application of the enhancements methods. For subjects 1-3 the ML-TP distance shows much less variation when including the FBC. For all subjects also (CSD + enh) gives more stable results than just CSD. Moreover, in all cases the combination (CSD + enh + FBC) outperforms CSD and for all but subject 1 the combined method (CSD + enh + FBC) also gives better results than the enhancement or FBC individually. It should be remarked that higher up the graph indicates a larger resection if used for pre-surgical evaluation, which is not necessarily positive. However, we prefer to have a stable and reproducible method that can be used with a safety margin, then a method that is more conservative, but shows large variations. 

\begin{figure}[t!]
        \centering
                \includegraphics[width= \textwidth]{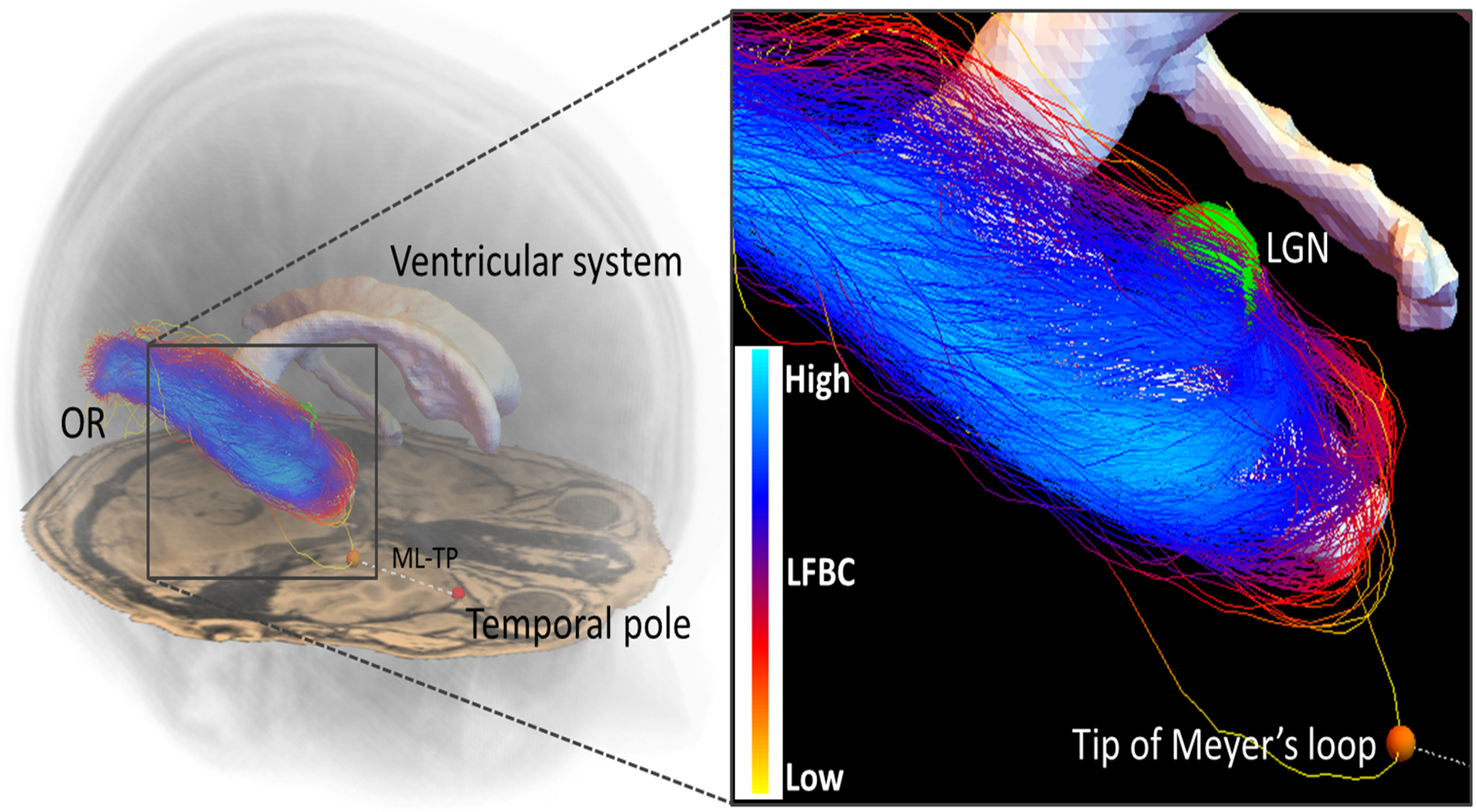}
        \caption{{\bf A reconstruction of the optic radiation and its positioning in the brain.}
        The left figure shows how the OR is positioned in the brain, the close-up on the right shows how the OR wraps around the ventricular system. The probabilistic tractography outputs many spurious fibers. The tip of the Meyer's loop, indicated by the orange sphere, is localized on a spurious fiber and is therefore very dependent on the realization of the tractography. As a result, the distance from the Meyer's loop to the Temporal pole (ML-TP) that is used in temporal lobe resection surgery, shows a high variation among different tractography outcomes.}\label{fig:prtscrn_OR}
\end{figure}

\begin{figure}[t!]
        \centering
                \includegraphics[width= \textwidth]{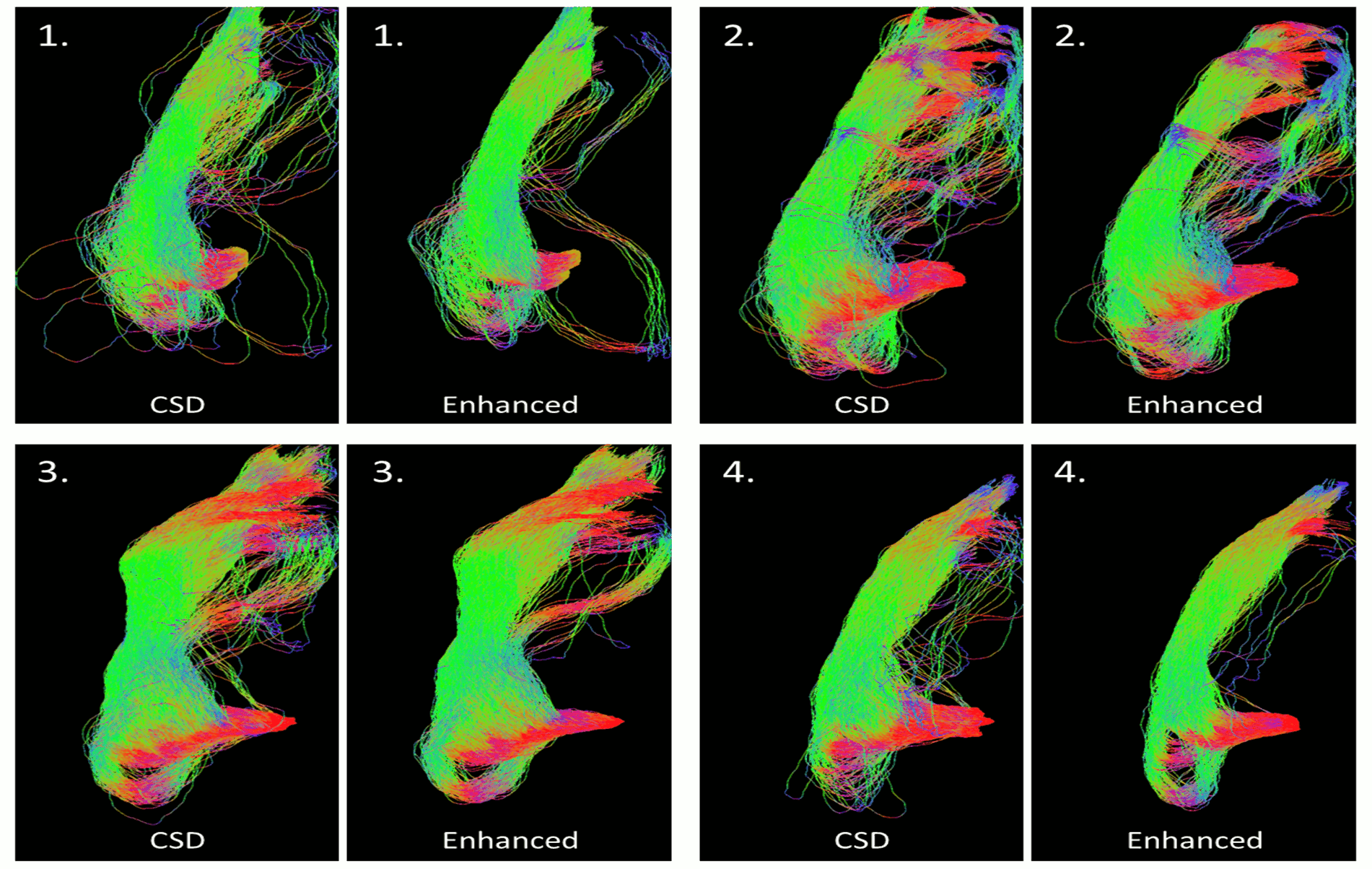}
        \caption{{\bf Reconstructions of the optic radiation of four subjects with and without use of enhancements.} For all subjects, the left image shows the result on the original data, the right image shows the result on the enhanced FOD. The enhanced version generally gives less spurious fibers and has a more pronounced tip of the Meyer's loop.}\label{fig:OR4subjects}
\end{figure}

\begin{figure}[t!]
        \centering
                \includegraphics[width= \textwidth]{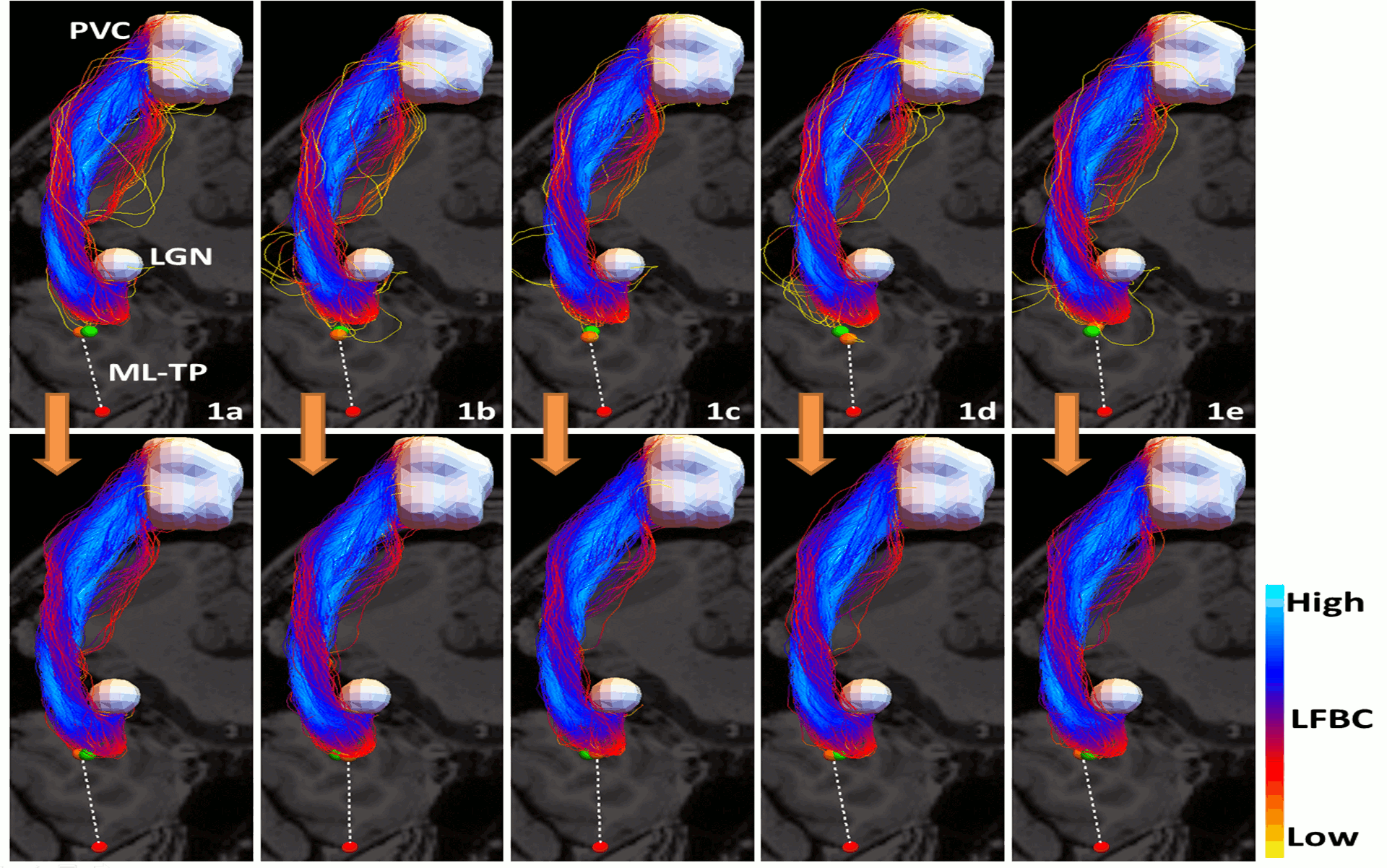}
        \caption{{\bf The effect of filtering spurious fibers from a probabilistic tractography on subject 1 in five different instances.}
        Top row: five different instances of the probabilistic tractography of the OR, viewed from the top, selecting only the 1000 most anterior fibers. Bottom row: the result after filtering the most spurious fibers for each of the instances. The red sphere indicates the temporal pole, the white volumes represent the LGN and the primary visual cortex. The orange spheres are the positions with minimal ML-TP distance. The green sphere indicates the position of the tip averaged over the five tractography results, before (top) or after filtering (bottom). There is less variation in the position of the tip of the Meyer's loop in the bottom row, i.e. after filtering, than in the top row. The fiber bundle in the left upper corner is the same as the one in Fig. \ref{fig:prtscrn_OR}.}\label{fig:ORcomparison}
\end{figure}

\section{Conclusions and Discussion}\label{se:conclusions}
We have proposed two new tools to improve alignment of fibers in tractography results: (A) the combination of CSD with contextual PDE enhancements and (B) a fiber to bundle coherence measure to classify spurious fibers. Both approaches rely on the same contextual processing via PDEs on the space of coupled positions and orientations. We validate our methodology with a variety of experiments on synthetic and human data.

In the first experiment we consider a digital phantom \cite{Daducci2013} that simulates DW-MRI data of a challenging configuration of multiple neural-like fiber bundles for different noise levels, see Fig. \ref{fig:isbifull}. The combination of CSD with enhancements and subsequent deterministic tracking was extensively tested for varying enhancement parameters, see Fig. \ref{fig:isbiresults}. The enhanced FOD peaks were compared with the ground truth fiber orientations, showing for all SNRs that the maxima of the enhanced FOD coincide better with the ground truth peaks than without application of enhancement. Also, this improvement is particularly high for very low SNR values. To quantitatively evaluate the impact of the enhancement on the tractographies we used the Tractometer evaluation system \cite{Cote2013}. The results, shown in Fig. \ref{fig:isbiresults} confirm the benefit, for all the metrics considered, of including the enhancement. Also an improved stability of the metrics with respect to different enhancement parameters is observed. Furthermore, we found that data with a lower SNR requires more regularization, obtained by choosing a higher diffusion time $t$ in the enhancement. These quantitative evaluations of local and global metrics are supported by the qualitative results in Figs. \ref{fig:isbifull} and \ref{fig:prtscrns}, where we saw that after enhancement fibers are better aligned and propagate better through crossings.

The second experiment is performed on human data of a representative area of the brain with crossing fiber bundles. We evaluate our combination of CSD and enhancement for three different (single-shell) acquisition protocols, corresponding to different $b$-values and number of gradient directions. We observed, see Fig. \ref{fig:trackingcomparison}, that whereas tractography on CSD without enhancement showed notable differences between the three acquisition protocols, tractography after our enhancement lead to a qualitatively similar reconstruction in all cases. This implies that the application of enhancement in the processing pipeline makes the tractography results less dependent on the scanning protocol used. 

We use the same dataset and the phantom dataset to compare our method qualitatively and quantitatively with previous work \cite{Duits2013,Tax2014} in which sharpening methods and nonlinear enhancement PDEs are applied to DTI. We observed qualitatively on real data in Fig. \ref{fig:glyphcomparison} and quantitatively in Table \ref{tab:isbiresults} the advantage of CSD, that allows to use linear enhancements with less extreme regularization parameters than with the DTI based method, resulting in a more reliable tractography.

For our second approach to improve fiber alignment, we introduced a fiber to bundle coherence measure that can be used for detecting and filtering spurious fibers. The fiber to bundle coherence (FBC) is computed from a tractography based density that we constructed using the same PDE foundation as in the first method. As an application we considered the reconstruction of the optic radiation, a fiber bundle of which the position of the anterior extent (the Meyer's loop) is of interest for temporal lobe resection surgery. Accurate and stable localization of the tip of the Meyer's loop is difficult due to the presence of spurious fibers, as shown in Fig. \ref{fig:prtscrn_OR}. We demonstrated in Figs. \ref{fig:OR4subjects}, \ref{fig:ORcomparison} and \ref{fig:MLTP_distances} that either by enhancement of the CSD FOD, or by removing the most spurious fibers using the FBC measure leads to a robust probabilistic tractography. In particular, the combination of both methods in one pipeline allows for a more stable localization of the tip of the Meyer's loop and a more stable determination of the Meyer's loop to Temporal Pole distance.

Our experiments show that our PDE enhancement methods for contextual processing are an effective and widely applicable tool to both enhance CSD data and to remove spurious fibers from tractographies. While we used CSD to construct an FOD, the PDE enhancement can be applied to an FOD obtained with any other method. We have seen that both our methods improve fiber alignment in tractography results and hence provide information on structural connectivity of the brain white matter more robustly. In the future, we aim to improve this framework by using data-adaptive smoothing, for example using local gauge frames \cite{Duits2015}.

\section*{Acknowledgments}
 We would like to thank Dr. A. Roebroeck from the Faculty of Psychology \& Neuroscience, Maastricht University for providing us with the human data used in Section \ref{se:evaluationDWMRI}. The study was approved by the local ethical commitee of Maastricht University. We gratefully acknowledge Academic Center for Epileptology Kempenhaeghe \& Maastricht UMC+ for providing us with the healthy volunteer data used in Section \ref{se:ORexperiment}. The study was approved by the Medical Ethics Committee of Maastricht University Medical Center (N 43386.068). Informed written consent was obtained from all subjects. The research leading to the results of this paper has received funding from the European Research Council under the European Community’s 7th Framework Programme (FP7/2007–2014)/ERC grant agreement No. 335555.
\nolinenumbers


\begin{thebibliography}{10}


\bibitem{leBihan1986}
Le~Bihan D, Breton E, Lallemand D, Grenier P, Cabanis E, et~al. (1986) {MR}
  imaging of intravoxel incoherent motions: application to diffusion and
  perfusion in neurologic disorders.
\newblock Radiology 161: 401--407.

\bibitem{Basser1994}
Basser PJ, Mattiello J, Le~Bihan D (1994) {MR} diffusion tensor spectroscopy
  and imaging.
\newblock Biophysical journal 66: 259--267.

\bibitem{tournier2011diffusion}
Tournier JD, Mori S, Leemans A (2011) Diffusion tensor imaging and beyond.
\newblock Magnetic Resonance in Medicine 65: 1532--1556.

\bibitem{jeurissen2013investigating}
Jeurissen B, Leemans A, Tournier JD, Jones DK, Sijbers J (2013) Investigating
  the prevalence of complex fiber configurations in white matter tissue with
  diffusion magnetic resonance imaging.
\newblock Human Brain Mapping 34: 2747--2766.

\bibitem{Descoteaux2012}
Descoteaux M, Poupon C (2012) Diffusion-weighted {MRI}.
\newblock In: Comprehensive Biomedical Physics. Elsevier.

\bibitem{Tournier2012}
Tournier JD, Calamante F, Connelly A (2012) {MR}trix: Diffusion tractography in
  crossing fiber regions.
\newblock International Journal of Imaging Systems and Technology 22: 53-66.

\bibitem{Jones2010}
Jones DK, Cercignani M (2010) Twenty-five pitfalls in the analysis of diffusion
  {MRI} data.
\newblock NMR in Biomedicine 23: 803--820.

\bibitem{Wiest2007}
Wiest-Daessl\'{e} N, Prima S, Coup\'{e} P, Morrissey SP, Barillot C (2007)
  Non-local means variants for denoising of diffusion-weighted and diffusion
  tensor {MRI}.
\newblock In: MICCAI (2). Springer, volume 4792 of \emph{Lecture Notes in
  Computer Science}, pp. 344-351.

\bibitem{Coupe2008}
Coup\'{e} P, Yger P, Prima S, Hellier P, Kervrann C, et~al. (2008) An optimized
  blockwise nonlocal means denoising filter for 3-{D} magnetic resonance
  images.
\newblock IEEE Transactions in Medical Imaging 27: 425-441.

\bibitem{Descoteaux2008}
Descoteaux M, Wiest-Daessl\'{e} N, Prima S, Barillot C, Deriche R (2008) Impact
  of {R}ician adapted non-local means filtering on {HARDI}.
\newblock In: MICCAI (2). Springer, volume 5242 of \emph{Lecture Notes in
  Computer Science}, pp. 122-130.

\bibitem{Poupon2000}
Poupon C, Clark CA, Frouin V, R\'egis J, Bloch I, et~al. (2000) Regularization
  of diffusion-based direction maps for the tracking of brain white matter
  fascicles.
\newblock {{N}euro{I}mage} 12: 184--195.

\bibitem{Coulon2001}
Coulon O, Alexander DC, Arridge SR (2001) A regularization scheme for diffusion
  tensor magnetic resonance images.
\newblock In: IPMI. Springer, volume 2082 of \emph{Lecture Notes in Computer
  Science}, pp. 92-105.

\bibitem{tschumperle2001}
Tschumperl{\'e} D, Deriche R (2001) Diffusion tensor regularization with
  constraints preservation.
\newblock In: Computer Vision and Pattern Recognition, 2001. CVPR 2001.
  Proceedings of the 2001 IEEE Computer Society Conference on. IEEE, volume~1,
  pp. I--948.

\bibitem{Burgeth2009a}
Burgeth B, Didas S, Weickert J (2009) A general structure tensor concept and
  coherence-enhancing diffusion filtering for matrix fields.
\newblock In: Visualization and Processing of Tensor Fields, Springer Berlin
  Heidelberg, Mathematics and Visualization. pp. 305-323.

\bibitem{Burgeth2009b}
Burgeth B, Breuß M, Pizarro L, Weickert J (2009) {PDE}-driven adaptive
  morphology for matrix fields.
\newblock In: SSVM. Springer, volume 5567 of \emph{Lecture Notes in Computer
  Science}, pp. 247-258.

\bibitem{Tuch2004}
Tuch DS (2004) Q-ball imaging.
\newblock Magnetic Resonance in Medicine 52: 1358--1372.

\bibitem{Descoteaux2007}
Descoteaux M, Angelino E, Fitzgibbons S, Deriche R (2007) Regularized, fast,
  and robust analytical {Q}-ball imaging.
\newblock Magnetic Resonance in Medicine 58: 497--510.

\bibitem{descoteaux2009deterministic}
Descoteaux M, Deriche R, Knosche T, Anwander A (2009) Deterministic and
  probabilistic tractography based on complex fibre orientation distributions.
\newblock Medical Imaging, IEEE Transactions on 28: 269--286.

\bibitem{Barmpoutis2008}
Barmpoutis A, Vemuri BC, Howland D, Forder JR (2008) Extracting tractosemas
  from a displacement probability field for tractography in {DW-MRI}.
\newblock In: MICCAI (1). Springer, volume 5241 of \emph{Lecture Notes in
  Computer Science}, pp. 9-16.

\bibitem{Goh2009}
Goh A, Lenglet C, Thompson PM, Vidal R (2009) Estimating orientation
  distribution functions with probability density constraints and spatial
  regularity.
\newblock In: MICCAI (1). Springer, volume 5761 of \emph{Lecture Notes in
  Computer Science}, pp. 877-885.

\bibitem{schultz2012}
Schultz T (2012) Towards resolving fiber crossings with higher order tensor
  inpainting.
\newblock In: New Developments in the Visualization and Processing of Tensor
  Fields, Springer. pp. 253--265.

\bibitem{Reisert2011}
Reisert M, Kiselev VG (2011) Fiber continuity: An anisotropic prior for {ODF}
  estimation.
\newblock IEEE Transactions on Medical Imaging 30: 1274-1283.

\bibitem{Tax2014}
Tax C, Duits R, Vilanova A, ter Haar~Romeny B, Hofman P, et~al. (2014)
  Evaluating contextual processing in diffusion {MRI}: Application to {O}ptic
  {R}adiation reconstruction for epilepsy surgery.
\newblock PLoS One .

\bibitem{Franken2008}
Franken E (2008) Enhancement of crossing elongated structures in images.
\newblock Ph.D. thesis, Eindhoven University of Technology, Department of
  Biomedical Engineering, The Netherlands.

\bibitem{Duits2011}
Duits R, Franken E (2011) Left-invariant diffusions on the space of positions
  and orientations and their application to crossing-preserving smoothing of
  {HARDI} images.
\newblock International Journal of Computer Vision 92: 231-264.

\bibitem{Creusen2011}
Creusen EJ, Duits R, Dela~Haije TCJ (2011) Numerical schemes for linear and
  non-linear enhancement of {DW-MRI}.
\newblock In: SSVM. Springer, volume 6667 of \emph{Lecture Notes in Computer
  Science}, pp. 14-25.

\bibitem{Duits2013}
Duits R, Dela~Haije TCJ, Creusen EJ, Ghosh A (2013) Morphological and linear
  scale spaces for fiber enhancement in {DW-MRI}.
\newblock Journal of Mathematical Imaging and Vision 46: 326-368.

\bibitem{franken2009crossing}
Franken E, Duits R (2009) Crossing-preserving coherence-enhancing diffusion on
  invertible orientation scores.
\newblock International Journal of Computer Vision 85: 253--278.

\bibitem{Mumford1994}
Mumford D (1994) Elastica and computer vision.
\newblock In: Algebraic Geometry and its Applications, Springer New York. pp.
  491-506.

\bibitem{Zweck00euclideangroup}
Zweck JW, Williams LR (2000) Euclidean group invariant computation of
  stochastic completion fields using shiftable-twistable functions.
\newblock In: Journal of Mathematical Imaging and Vision. pp. 100--116.

\bibitem{SanguinettiJOV2010}
Sanguinetti G, Citti G, Sarti A (2010) A model of natural image edge
  co-occurrence in the rototranslation group.
\newblock Journal of Vision 10.

\bibitem{August2003}
August J, Zucker SW (2003) Sketches with curvature: The curve indicator random
  field and {M}arkov processes.
\newblock IEEE Transactions on Pattern Analysis and Machine Intelligence 25:
  387-400.

\bibitem{Duits2008ESL}
Duits R, van Almsick M (2008) {The explicit solutions of linear left-invariant
  second order stochastic evolution equations on the 2D Euclidean motion
  group}.
\newblock Quarterly of Applied Mathematics 66: 27--67.

\bibitem{Momayyez2013}
MomayyezSiahkal P, Siddiqi K (2013) 3{D} stochastic completion fields for
  mapping connectivity in diffusion {MRI}.
\newblock IEEE Transactions on Pattern Analysis and Machine Intelligence 35:
  983-995.

\bibitem{Citti2006}
Citti G, Sarti A (2006) A cortical based model of perceptual completion in the
  roto-translation space.
\newblock Journal of Mathematical Imaging and Vision 24: 307-326.

\bibitem{Duits2010LIP1}
Duits R, Franken E (2010) {Left-invariant parabolic evolutions on SE(2) and
  contour enhancement via invertible orientation scores Part I: Linear
  left-invariant diffusion equations on SE(2)}.
\newblock Quarterly of Applied Mathematics 68: 255--292.

\bibitem{Agrachev2009}
Agrachev A, Boscain U, Gauthier JP, Rossi F (2009) The intrinsic hypoelliptic
  laplacian and its heat kernel on unimodular {L}ie groups.
\newblock Journal of Functional Analysis 256: 2621 - 2655.

\bibitem{Vesna2010}
Prčkovska V, Rodrigues P, Duits R, Haar~Romenij Bt, Vilanova A (2010)
  Extrapolating fiber crossings from {DTI} data: can we infer similar fiber
  crossings as in {HARDI}?
\newblock In: MICCAI. Workshop on Computational Diffusion MRI.

\bibitem{Vesna2015}
Prčkovska V, Andorr\`a M, Villoslada P, Martinez-Heras E, Duits R, et~al.
  (2015) Contextual diffusion image post-processing aids clinical applications.
\newblock In: Visualization and Processing of Tensors and Higher Order
  Descriptors for Multi-Valued Data. Springer.

\bibitem{Reisert2012}
Reisert M, Skibbe H (2012) Left-invariant diffusion on the motion group in
  terms of the irreducible representations of {SO}(3).
\newblock Computing Research Repository abs/1202.5414.

\bibitem{Reisert2013}
Reisert M, Skibbe H (2013) Fiber continuity based spherical deconvolution in
  spherical harmonic domain.
\newblock In: Mori K, Sakuma I, Sato Y, Barillot C, Navab N, editors, MICCAI
  (3). Springer, volume 8151 of \emph{Lecture Notes in Computer Science}, pp.
  493-500.

\bibitem{DelaHaije2014}
Dela~Haije T, Duits R, Tax C (2014) Sharpening fibers in diffusion weighted
  {MRI} via erosion.
\newblock In: Westin C, Vilanova A, Burgeth B, editors, Visualization and
  processing of tensors and higher order descriptors for multi-valued data,
  Mathematics and visualization.

\bibitem{Tournier2008}
Tournier JD, Yeh CH, Calamante F, Cho KH, Connelly A, et~al. (2008) Resolving
  crossing fibres using constrained spherical deconvolution: Validation using
  diffusion-weighted imaging phantom data.
\newblock NeuroImage 42: 617-625.

\bibitem{Tax2013}
Tax C, Jeurissen B, Vos S, Viergever M, Leemans A (2014) Recursive calibration
  of the fiber response function for spherical deconvolution of diffusion {MRI}
  data.
\newblock NeuroImage 86: 67--80.

\bibitem{schultz2013auto}
Schultz T, Groeschel S (2013) Auto-calibrating spherical deconvolution based on
  {ODF} sparsity.
\newblock In: MICCAI 2013, Springer. pp. 663--670.

\bibitem{Jeurissen2014411}
Jeurissen B, Tournier JD, Dhollander T, Connelly A, Sijbers J (2014)
  Multi-tissue constrained spherical deconvolution for improved analysis of
  multi-shell diffusion {MRI} data.
\newblock NeuroImage 103: 411 - 426.

\bibitem{10.3389/fninf.2014.00028}
Roine T, Jeurissen B, Perrone D, Aelterman J, Leemans A, et~al. (2014)
  Isotropic non-white matter partial volume effects in constrained spherical
  deconvolution.
\newblock Frontiers in Neuroinformatics 8.

\bibitem{Roine2015}
Roine T, Jeurissen B, Perrone D, Aelterman J, Philips W, et~al. (2015) Informed
  constrained spherical deconvolution (i{CSD}).
\newblock Medical Image Analysis : In press.

\bibitem{calamante2010track}
Calamante F, Tournier JD, Jackson GD, Connelly A (2010) Track-density imaging
  ({TDI}): super-resolution white matter imaging using whole-brain
  track-density mapping.
\newblock Neuroimage 53: 1233--1243.

\bibitem{dhollander2014track}
Dhollander T, Emsell L, Van~Hecke W, Maes F, Sunaert S, et~al. (2014) Track
  orientation density imaging ({TODI}) and track orientation distribution
  ({TOD}) based tractography.
\newblock NeuroImage 94: 312--336.

\bibitem{cote2012tractometer}
C{\^o}t{\'e} MA, Bor{\'e} A, Girard G, Houde JC, Descoteaux M (2012)
  Tractometer: online evaluation system for tractography.
\newblock In: Medical Image Computing and Computer-Assisted
  Intervention--MICCAI 2012, Springer. pp. 699--706.

\bibitem{Cote2013}
C\^ot\'e MA, Girard G, Boré A, Garyfallidis E, Houde JC, et~al. (2013)
  Tractometer: Towards validation of tractography pipelines.
\newblock Medical Image Analysis 17: 844-857.

\bibitem{Daducci2014}
Daducci A, Canales-Rodr\'iguez EJ, Descoteaux M, Garyfallidis E, Gur Y, et~al.
  (2014) Quantitative comparison of reconstruction methods for intra-voxel
  fiber recovery from diffusion {MRI}.
\newblock {IEEE} {T}ransactions on {M}edical {I}maging 33: 384--399.

\bibitem{falconer1963follow}
Falconer MA, Serafetinides EA (1963) A follow-up study of surgery in temporal
  lobe epilepsy.
\newblock Journal of neurology, neurosurgery, and psychiatry 26: 154.

\bibitem{powell2005mr}
Powell H, Parker G, Alexander D, Symms M, Boulby P, et~al. (2005) {MR}
  tractography predicts visual field defects following temporal lobe resection.
\newblock Neurology 65: 596--599.

\bibitem{Sherbondy2008Contrack}
Sherbondy AJ, Dougherty RF, Ben-Shachar M, Napel S, Wandell BA (2008) Contrack:
  finding the most likely pathways between brain regions using diffusion
  tractography.
\newblock Journal of Vision 8: 15.1—16.

\bibitem{Meesters2013}
Meesters S (2013) Diffusion weighted tractography to reconstruct the optic
  radiation in support of temporal lobe epilepsy surgery.
\newblock Master's thesis, Eindhoven University of Technology.
\newblock \url{http://repository.tue.nl/789374}.

\bibitem{Tournier2004}
Tournier JD, Calamante F, Gadian DG, Connelly A (2004) Direct estimation of the
  fiber orientation density function from diffusion-weighted {MRI} data using
  spherical deconvolution.
\newblock NeuroImage 23: 1176 - 1185.

\bibitem{Tournier2007}
Tournier JD, Calamante F, Connelly A (2007) Robust determination of the fibre
  orientation distribution in diffusion {MRI}: Non-negativity constrained
  super-resolved spherical deconvolution.
\newblock NeuroImage 35: 1459 - 1472.

\bibitem{Driscoll1994S2conv}
Driscoll J, Healy D (1994) Computing {F}ourier transforms and convolutions on
  the 2-sphere.
\newblock Advances in Applied Mathematics 15: 202 - 250.

\bibitem{Tournier2013}
Tournier JD, Calamante F, Connelly A (2013) Determination of the appropriate b
  value and number of gradient directions for high-angular-resolution
  diffusion-weighted imaging.
\newblock NMR in Biomedicine .

\bibitem{Duits2007}
Duits R, Burgeth B (2007) Scale spaces on {L}ie groups.
\newblock In: SSVM. Springer, volume 4485 of \emph{Lecture Notes in Computer
  Science}, pp. 300-312.

\bibitem{Creusen2013}
Creusen E, Duits R, Vilanova A, Florack L (2013) Numerical schemes for linear
  and non-linear enhancement of {DW-MRI}.
\newblock Numerical Mathematics: Theory, Methods and Applications 6: 326-368.

\bibitem{rodrigues2010}
Rodrigues P, Duits R, ter Haar~Romeny BM, Vilanova A (2010) Accelerated
  diffusion operators for enhancing {DW}-{MRI}.
\newblock In: Proc. of the 2nd EG conference on VCBM. Eurographics Association,
  pp. 49--56.

\bibitem{Hormander1967}
H\"ormander L (1967) Hypoelliptic second order differential equations.
\newblock Acta Mathematica 119: 147-171.

\bibitem{Daducci2013}
Daducci A, Caruyer E, Descoteaux M, Thiran JP (2013).
\newblock {HARDI} reconstruction challenge.
\newblock IEEE International Symposium on Biomedical Imaging.
\newblock \url{http://hardi.epfl.ch/statis/events/2013_ISBI/}.

\bibitem{Mathematica10}
{Wolfram Research Inc} (2014) Mathematica.
\newblock Wolfram Research Inc., 10.0 edition.

\bibitem{chamberland2014real}
Chamberland M, Whittingstall K, Fortin D, Mathieu D, Descoteaux M (2014)
  Real-time multi-peak tractography for instantaneous connectivity display.
\newblock Frontiers in neuroinformatics 8.

\bibitem{Caruyer2014}
Caruyer E, Daducci A, Descoteaux M, Houde JC, Thiran JP, et~al. (2014)
  Phantomas: a flexible software library to simulate diffusion {MR} phantoms.
\newblock International Society for Magnetic Resonance in Medicine .

\bibitem{Close2009}
Close TG, Tournier JD, Calamante F, Johnston LA, Mareels I, et~al. (2009) A
  software tool to generate simulated white matter structures for the
  assessment of fibre-tracking algorithms.
\newblock NeuroImage 47: 1288 - 1300.

\bibitem{Assaf2005}
Assaf Y, Basser PJ (2005) {Composite hindered and restricted model of diffusion
  (CHARMED) MR imaging of the human brain}.
\newblock NeuroImage 27: 48--58.

\bibitem{SODERMAN199594}
S\"oderman O, J\"onsson B (1995) Restricted diffusion in cylindrical geometry.
\newblock Journal of Magnetic Resonance, Series A 117: 94 - 97.

\bibitem{Girard2014}
Girard G, Whittingstall K, Deriche R, Descoteaux M (2014) Towards quantitative
  connectivity analysis: reducing tractography biases.
\newblock NeuroImage 98: 266 - 278.

\bibitem{smith2013sift}
Smith RE, Tournier JD, Calamante F, Connelly A (2013) {SIFT}:
  spherical-deconvolution informed filtering of tractograms.
\newblock Neuroimage 67: 298--312.

\bibitem{smith2012anatomically}
Smith RE, Tournier JD, Calamante F, Connelly A (2012) Anatomically-constrained
  tractography: improved diffusion {MRI} streamlines tractography through
  effective use of anatomical information.
\newblock Neuroimage 62: 1924--1938.

\bibitem{aganj2011hough}
Aganj I, Lenglet C, Jahanshad N, Yacoub E, Harel N, et~al. (2011) A {H}ough
  transform global probabilistic approach to multiple-subject diffusion {MRI}
  tractography.
\newblock Medical image analysis 15: 414--425.

\bibitem{nilsson2007intersubject}
Nilsson D, Starck G, Ljungberg M, Ribbelin S, J{\"o}nsson L, et~al. (2007)
  Intersubject variability in the anterior extent of the optic radiation
  assessed by tractography.
\newblock Epilepsy research 77: 11--16.

\bibitem{Duits2015}
Duits R, Janssen M, Hannink J, Sanguinetti G (2015) Locally adaptive frames in
  the roto-translation group and their applications in medical imaging.
\newblock arXiv:{150208002v4} .

\end{thebibliography}
\end{document}